\documentclass[preprint,12pt]{elsarticle}

\usepackage{amssymb}
\usepackage{amsmath}
\usepackage{hyperref}       % hyperlinks
\usepackage{url}            % simple URL typesetting
\usepackage{booktabs}       % professional-quality tables
\usepackage{amsfonts}       % blackboard math symbols
\usepackage{nicefrac}       % compact symbols for 1/2, etc.
\usepackage{microtype}      % microtypography
\usepackage{xcolor}         % colors
\usepackage{multirow}
\usepackage{CJKutf8}
\usepackage{array}      % for >{\centering\arraybackslash}

\newtheorem{definition}{Definition}

\journal{Expert Systems with Applications}

\begin{document}

\begin{frontmatter}



\title{Understanding InfoNCE: Transition Probability Matrix Induced Feature Clustering}


% \author{} %% Author name

% \affiliation{organization={},%Department and Organization

\author[inst1]{Ge Cheng}
\ead{chengge@xtu.edu.cn}

\author[inst2]{Shuo Wang \corref{cor1}}
\ead{202331630305@smail.xtu.edu.cn}

\author[inst3]{Yun Zhang}
\ead{zhangyun@hnpa.edu.cn}

% \author[inst1,inst3]{Ge Cheng\corref{cor1}}  % 多单位的作者

\cortext[cor1]{Corresponding author}  % 通讯作者说明

% 单位1
\affiliation[inst1]{organization={School of Mathematics and Computational Science, Xiangtan University},
            addressline={Yuhu District}, 
            city={Xiangtan},
            postcode={411105}, 
            state={Hunan Province},
            country={China}}

% 单位2
\affiliation[inst2]{organization={School of Computer Science \& School of Cyberspace Science, Xiangtan University},
            addressline={Yuhu District}, 
            city={Xiangtan},
            postcode={411105}, 
            state={Hunan Province},
            country={China}}
            
\affiliation[inst3]{organization={Investigation Department, Hunan Police Academy},
            addressline={Xingsha District}, 
            city={Changsha},
            postcode={410138}, 
            state={Hunan Province},
            country={China}}

% % 单位3
% \affiliation[inst3]{organization={Department of Computer Science, Stanford University},
%             addressline={450 Serra Mall}, 
%             city={Stanford},
%             postcode={94305}, 
%             state={CA},
%             country={United States}}

%% Abstract
\begin{abstract}
Contrastive learning has emerged as a cornerstone of unsupervised representation learning across vision, language, and graph domains, with InfoNCE as its dominant objective. Despite its empirical success, the theoretical underpinnings of InfoNCE remain limited. In this work, we introduce an explicit feature space to model augmented views of samples and a transition probability matrix to capture data augmentation dynamics. We demonstrate that InfoNCE optimizes the probability of two views sharing the same source toward a constant target defined by this matrix, naturally inducing feature clustering in the representation space. Leveraging this insight, we propose Scaled Convergence InfoNCE (SC-InfoNCE), a novel loss function that introduces a tunable convergence target to flexibly control feature similarity alignment. By scaling the target matrix, SC-InfoNCE enables flexible control over feature similarity alignment, allowing the training objective to better match the statistical properties of downstream data.  Experiments on benchmark datasets, including image, graph, and text tasks, show that SC-InfoNCE consistently achieves strong and reliable performance across diverse domains.

\end{abstract}


\begin{highlights}
\item Introduce explicit feature space and transition probability matrix for augmentations.
\item Show InfoNCE drives co-occurrence probability to a constant convergence target.
% \item Derive SCL gradients from transition covariance, revealing clustering effects.
\item Propose SC-InfoNCE with tunable delta and gamma to modulate the target.
\item Achieve consistent gains on vision, graph, and text benchmarks over baselines.
\end{highlights}

%% Keywords

\end{frontmatter}


\section{Introduction}

Contrastive learning (CL) has become a pivotal method for unsupervised representation learning, delivering impressive results in computer vision \cite{byol,ref2,ref4}, natural language processing \cite{simcse,llmc2024}, and graph representation learning \cite{surveygcl,ref5}. Its core idea is to maximize the similarity between positive pairs, which are defined as two augmented views of the same instance, and to minimize the similarity between negative pairs, which are views of different instances. This strategy enables the model to learn rich representations without requiring manual labels. Among the various CL methods, SimCLR \cite{simclr} and MoCo \cite{moco,fastmoco} stand out, both relying on the widely adopted InfoNCE loss \cite{infonce}.

Despite InfoNCE’s empirical success, its theoretical foundations remain underexplored \cite{basetheo,kernelana}. Prior studies have interpreted InfoNCE as (i) a lower bound on mutual information \cite{ref18,ref16,ref17}, (ii) a variational objective for maximizing Kullback–Leibler divergence \cite{fmicl}, or (iii) an implicit spectral clustering mechanism on similarity graphs \cite{Spectral}. These perspectives offer valuable insights, but none fully explain or predict how InfoNCE shapes the organization of the representation space.

In this work, we address this question by introducing an \emph{explicit feature space} to represent augmented views of samples, along with a \emph{transition probability matrix} that captures the dynamics induced by data augmentation. Specifically, we consider a finite set of \emph{non-augmentable features} \(S\), and define an augmentation distribution \(T\). Applying \(T\) to each element in \(S\) yields a feature space of augmented views, whose transformation behavior is modeled by a transition probability matrix \(A\). Each entry \(A_{ij}\) denotes the probability that a view generated from source \(i\) is observed as feature \(j\) (see Section~\ref{sec2.3}). This construction enables us to track the evolution of individual features and uncover how augmentation patterns influence the dynamics of representation learning.

\begin{table}[t]
  \centering
  \caption{Comparison between the predicted (Independent Sampling) and empirical measured probability $\mathbb{P}$ that samples from different categories are assigned to the same latent distribution under a given transition probability matrix (TPM). The last column reports the mean absolute error (MAE) between predicted and empirical values for each case.}
  \label{tab:stm_cmp}
  \resizebox{\textwidth}{!}{%
  \begin{tabular}{cccccccccc}
    \toprule
    \multicolumn{3}{c}{\multirow{2}{*}{\textbf{TPM}}} &
    \multicolumn{3}{c}{\textbf{Predicted $\mathbb{P}_{ij}$}} &
    \multicolumn{3}{c}{\textbf{Measured $\mathbb{P}_{ij}$ (mean ± SD)}} &
    \multirow{2}{*}{\textbf{MAE}} \\
    \cmidrule(r){4-6} \cmidrule(r){7-9}
    & & & $j=1$ & $j=2$ & $j=3$ & $j=1$ & $j=2$ & $j=3$ & \\
    \midrule
    0.5 & 0.3 & 0.2 &
    0.00122 & 0.00094 & 0.00087 &
    0.00115{\tiny$\pm$0.00001} & 0.00096{\tiny$\pm$0.00000} & 0.00089{\tiny$\pm$0.00001} &
    0.00003 \\
    0.2 & 0.5 & 0.3 &
    0.00094 & 0.00107 & 0.00098 &
    0.00092{\tiny$\pm$0.00001} & 0.00110{\tiny$\pm$0.00001} & 0.00098{\tiny$\pm$0.00001} &
    0.00002 \\
    0.2 & 0.3 & 0.5 &
    0.00087 & 0.00098 & 0.00114 &
    0.00087{\tiny$\pm$0.00000} & 0.00100{\tiny$\pm$0.00000} & 0.00113{\tiny$\pm$0.00001} &
    0.00001 \\
    \bottomrule
  \end{tabular}
  }
\end{table}

To disentangle the role of InfoNCE, we first examine a \emph{Simplified Contrastive Loss} (SCL), which retains the mechanism of pulling positive pairs together and pushing apart negatives, but omits the temperature-scaled softmax. We find that the \emph{gradient direction} is determined by the covariance of a transition probability vector derived from the transition probability matrix \(A\), thereby inducing a clustering effect based on directional differences. However, while SCL yields well-formed clusters, it lacks a mechanism to regulate their relative positioning, resulting in uncontrolled inter-cluster alignment. In contrast, InfoNCE dynamically reshapes the training objective during learning, adapting to the evolving structure of the data. It drives the empirical co-occurrence probability \(\mathbb{P}_{ij}\), defined as the probability that two views originate from the same latent source, to converge to a constant value determined by the dataset and augmentation distribution. Empirical analysis (see Table~\ref{tab:stm_cmp}) shows a close match between predicted and observed values of \(\mathbb{P}_{ij}\), validating our theoretical predictions.

Building on this insight, we introduce \emph{Scaled-Convergence InfoNCE} (SC-InfoNCE), a novel loss function designed to modulate the convergence target via two tunable parameters, \(\delta\) and \(\gamma\). Intuitively, \(\delta\) controls the scaling coefficient of the convergence target, while \(\gamma\) adjusts its bias. Together, they enable flexible control over the target scale. Our analysis shows that a smaller convergence target increases sensitivity to subtle differences, resulting in noisier and less discriminative representations, while a larger target enhances feature separation but may lead to representation collapse. Empirical results on benchmark datasets spanning vision, graph, and text domains demonstrate that appropriately scaling the convergence target leads to consistently better performance than existing contrastive learning loss functions. Our contributions are summarized as follows:
\begin{itemize}

  \item We introduce an explicit feature space and a transition probability matrix to systematically model the impact of data augmentation on representation learning dynamics, providing theoretical insights into the behavior of InfoNCE.
  \item We prove that InfoNCE optimizes the empirical co-occurrence probability \(\mathbb{P}_{ij}\) toward a constant convergence target, naturally inducing feature clustering.
  \item We introduce SC-InfoNCE, which adjusts the convergence target through two tunable parameters, providing control over its scale and improving performance across domains.
  \item By tuning the convergence target, SC-InfoNCE provides consistent and controllable performance across the evaluated benchmarks, often matching strong baselines and occasionally surpassing them.
\end{itemize}

\section{Background and Definitions}
\label{sec2}

In this section, we formalize the core concepts and notations used throughout the paper. Let \(C\) be a (possibly infinite) set of latent classes with prior distribution \(\rho\).
Each class \(c\in C\) induces a class‑conditional distribution \(D_c\) over the input space \(\mathcal{X}\).
The marginal data distribution over \(\mathcal{X}\) is then
\[
  D(x)\;=\;\sum_{c\in C}\rho(c)\,D_c(x),
\]
where $x$ represents a point in the input space  \(\mathcal{X}\). We construct an unlabeled dataset by drawing \(n\) i.i.d.\ samples from \(D\),
yielding \(\mathcal{D}=\{x_i\}_{i=1}^{n}\) with \(x_i\sim D\).

\subsection{Mutual Information}

Mutual information, a fundamental concept in information theory, quantifies the dependence between random variables. In the context of supervised learning, let \( X \) denote the input and \( Y \) the corresponding label. A neural network \( f(\cdot) \) maps \( X \) to a latent representation \( Z = f(X) \), which ideally should capture task-relevant features of \( X \) while discarding irrelevant details. The information bottleneck (IB) principle \cite{infobtn} formalizes this intuition by optimizing the trade-off:
\[
    \min_{Z} \bigl[ I(X;Z) \;-\; \beta \, I(Z;Y) \bigr],
\]
where \( I(X;Z) \) encourages compression of the input, and \( I(Z;Y) \) promotes predictive sufficiency. The parameter \( \beta \) controls the trade-off. In practice, the IB objective is approximated by minimizing the cross-entropy loss to maximize \( I(Z;Y) \), while implicitly reducing \( I(X;Z) \) through architectural bottlenecks and regularization.

\subsection{InfoNCE}
In unsupervised learning, labels $Y$ are unavailable. Noise Contrastive Estimation (NCE) addresses this by treating a positive sample as drawn from a target distribution, with negatives from a noise distribution. The model learns to discriminate between them, thereby acquiring representations. Formally, a transformation $t(\cdot)$ is applied to $X$ to obtain a correlated view $\tilde{X} = t(X)$, serving as a surrogate for $Y$, and yielding the objective:
\begin{equation}
    \max \; I\big(Z;\, f(\tilde{X})\big).
\end{equation}
This objective is commonly instantiated in contrastive learning, where an augmentation distribution $T$ defines transformations over inputs. For each $x$, two views $t_1(x)$ and $t_2(x)$ are sampled with $t_1, t_2 \sim T$, forming a positive pair. Other views in the batch act as negatives. Let $z_1 = f(t_1(x))$ and $z_2 = f(t_2(x))$. The model then maximizes mutual information between $z_1$ and $z_2$ is defined as:
\begin{align}
    I(z_1; z_2) &= \int p(z_1, z_2) \log \frac{p(z_1, z_2)}{p(z_1)p(z_2)} \, dz_1 dz_2 \nonumber \\
                &= \mathbb{E}_{p(z_1, z_2)} \left[ \log \frac{p(z_2 \mid z_1)}{p(z_2)} \right],
\end{align}
but the conditional distribution $p(z_2 \mid z_1)$ is intractable. We instead introduce a variational approximation $q(z_2 \mid z_1)$ and apply Jensen’s inequality:
\begin{align}
    I(z_1; z_2) &\geq \mathbb{E}_{p(z_1, z_2)} \left[ \log \frac{q(z_2 \mid z_1)}{p(z_2)} \right] \nonumber \\
                &= \mathbb{E}_{p(z_1, z_2)} \left[ \log \frac{\exp\left(\text{sim}(z_1, z_2)/\tau\right)}{\sum_{k=1}^{N} \exp\left(\text{sim}(z_1, z_k)/\tau\right)} \right] + \log N \nonumber \\
                &= - \mathcal{L}_{\text{InfoNCE}} + \log N,
\label{eq3}
\end{align}
where $\text{sim}(\cdot)$ denotes a similarity function and $\tau$ is a temperature parameter. This objective corresponds to the InfoNCE loss \cite{infonce}, which provides a lower bound on the mutual information between augmented views.

\subsection{Explicit Feature Space}
\label{sec2.3}

Although the InfoNCE loss is grounded in mutual information theory, its empirical behavior remains not fully understood, particularly regarding how it influences feature clustering and whether theoretical performance bounds exist. To investigate this, we study the evolution of features under data augmentations and model updates. However, existing theoretical analyses often rely on latent features that are difficult to access in practice, limiting their applicability. 

In this section, we introduce an explicit feature space, defined over concrete and enumerable samples, which shifts the focus of contrastive loss analysis from abstract latent representations to the mechanisms by which it separates concrete examples. All formal notions below are illustrated on a graph dataset $\mathcal{G}=\{G_i\}_{i=1}^n$, where each graph $G=(V,E)$ has node features $X\!\in\!\mathbb{R}^{|V|\times d}$.

\begin{definition}[Closure]
Let $T$ be a distribution over graph augmentation functions. The \emph{closure} of a dataset $\mathcal{G}$ under $T$ is defined as
\[
\operatorname{cl}_T(\mathcal{G}) := \left\{\, t_k \circ \cdots \circ t_1(G)\;\middle|\; G \in \mathcal{G},\; k \ge 0,\; t_i \in T \right\},
\]
where the empty composition ($k = 0$) is defined as the identity, i.e., $t^{(0)}(G) = G$.
\end{definition}

\begin{definition}[Generating Set]
Given a dataset $\mathcal{G}$ and an augmentation distribution $T$, the \emph{generating set} is defined as
\[
S := \left\{\, G \in \mathcal{G} \;\middle|\; G \notin \operatorname{cl}_T(\mathcal{G} \setminus \{G\}) \right\}.
\]
Elements in $S$ are called \emph{base features}. If $S = \varnothing$, we define a \emph{zero-sample} $\boldsymbol{0}$ such that $\operatorname{cl}_T(\{\boldsymbol{0}\}) = \mathcal{G}$.
\end{definition}

\begin{definition}[Explicit Feature Space]
Given a dataset $\mathcal{G}$ and an augmentation distribution $T$, let $S \subseteq \mathcal{G}$ be a generating set. The pair $(S, T)$ defines the \emph{explicit feature space} over $(\mathcal{G}, T)$, consisting of all elements reachable from $S$ via finite compositions of augmentations in $T$, i.e., $\operatorname{cl}_T(S)$. Each element in this space is called an \emph{explicit feature}. For a base feature $G \in S$, we refer to $\operatorname{cl}_T(\{G\})$ as its \emph{sub-features}.
\end{definition}

The explicit feature space encompasses both the original dataset and all possible augmented views. In some cases, such as diffusion models \cite{ddpm,ldm}, where augmentation corresponds to iterative noise injection, the generating set becomes $S = {\boldsymbol{0}}$. In this case, the explicit feature space coincides with the model’s latent space.

\subsection{Contrastive Learning Process}
\label{sec2.4}
Before presenting our theory, we briefly outline a standard contrastive learning pipeline (e.g., SimCLR or GraphCL~\cite{graphcl}) to contextualize later probability computations:  
(i) Sample an unlabeled dataset $\mathcal{G} \sim \rho$ from data space $\mathcal{X}$.  
(ii) For a sample $G_1$, define an augmentation distribution $T$, sample $t_1, t_2 \sim T$ independently, and apply them to obtain views $\tilde{G}_1^{(1)} = t_1(G_1)$ and $\tilde{G}_1^{(2)} = t_2(G_1)$. Common augmentations include random crop, color jitter, or blur (images~\cite{ref6}); synonym replacement or masking (text~\cite{ref7}); and node/edge drop or attribute masking (graphs~\cite{ref8}).  
(iii) Use a neural network $f(\cdot)$ to embed the views: $z_1 = f(\tilde{G}_1^{(1)})$, $z_2 = f(\tilde{G}_1^{(2)})$.  
(iv) Train $f$ using the InfoNCE loss to learn meaningful representations.

\section{Analysis of Contrastive Learning}
\label{sec3}

In this section, we analyze CL from a gradient-based perspective and demonstrate how it captures distinct explicit features. We begin with the gradient of a SCL, extend the analysis to the widely used InfoNCE loss, and conclude with experimental validation.

\subsection{Gradient of SCL}\label{sec3.1}

The Simplified Contrastive Loss reflects the core objective of contrastive learning: aligning positive pairs while repelling negatives. It represents the most fundamental and canonical form of contrastive objectives. Let $z_1 = f(\tilde{G}_1^{(1)})$ and $z_2 = f(\tilde{G}_1^{(2)})$ be the embeddings of two augmented views of the same sample, each treated as an explicit feature. We generalize by using indices $i, j, k$ to denote explicit features, and write $z^+_i, z^+_j$ and $z^-_k$ for positive and negative examples, respectively. The loss is defined as:
\begin{equation}
    \mathcal{L}_{\mathrm{SCL}}(G) = -\,sim(z^{+}_{i}, z^{+}_{j}) \;+\; \lambda \sum^{Z} sim(z^{+}_{i}, z^{-}_{k})\,,
\end{equation}
where $G$ denotes a graph sample and $\lambda$ is a hyperparameter balancing the negative term.

Following the SimCLR training process in Section \ref{sec2.4}, suppose we have a graph dataset $\mathcal{G}$ with $n$ samples. Removing duplicates yields a deduplicated set $F$. Given a known augmentation distribution $T$, we define $F_+$ as the set of features not in $F$ but obtainable via augmentation. This allows us to define a feature sampling probability distribution $P$, where each feature in $F_+$ has zero sampling probability. When $n$ is sufficiently large, $F$, $F_+$, and $P$ together approximate the unlabeled data space $\mathcal{X}$ and its underlying distribution $\rho$. Let $A$ denote the transition probability matrix over features in $F \cup F_+$ induced by the augmentation distribution $T$, where $A_{i,j}$ denotes the probability that feature $j$ is generated by augmenting feature $i$, i.e., $j = T(i)$. As $T$ and $F \cup F_+$ are known, $A$ can be estimated. Assuming independent sampling of positive and negative pairs\footnote{In practical implementations such as SimCLR and GraphCL, the sampling of positive and negative pairs is not strictly independent. A more precise derivation that accounts for this dependency is given in Appendix~\ref{apd:the}.}, the probability of sampling feature $j$ as a positive for a given sample $G_i$ (corresponding to feature $i$) is $A_{i,j}$, and the probability of sampling feature $j$ as a negative is $P A_j$.

Denote by $\pi^p$ the probability matrix where each entry $\pi^p_{i,j}$ corresponds to the likelihood of selecting features $i$ and $j$ as a positive pair. The corresponding negative pair probabilities are given by the matrix $\pi^n$. Let $S$ be the similarity matrix between their hidden representations. Under the SCL objective, the expected gradient with respect to model parameters $\theta$ is given by:
\begin{align}
    \mathbb{E}[\nabla_\theta \mathcal{L}_{\mathrm{SCL}}] = (\pi^p - \lambda\,\pi^n) \nabla_\theta S = \pi\, \nabla_\theta S\,,
\end{align}
where $\pi$ encodes the effective gradient direction between features. Substituting the definitions and computing $\pi_{i,j}$ for any $i,j \in F \cup F_+$ yields:
\begin{align}
    \pi_{i,j} &= \sum_{k=1}^{|F|} P_k\Big(A_{k,i}A_{k,j} - \lambda A_{k,i} (P A)_j\Big) \nonumber \\
              &= \sum_{k=1}^{|F|} P_k A_{k,i} A_{k,j} - \lambda \sum_{k=1}^{|F|}(P_k A_{k,i}) \sum_{l=1}^{|F|}(P_l A_{l,j}) \nonumber \\
              &= \mathbb{E}[A_i A_j] - \lambda\, \mathbb{E}[A_i]\, \mathbb{E}[A_j]\,.
\label{eq6}
\end{align}

The entry $\pi_{i,j}$ reflects the expected gradient interaction between features $i$ and $j$, determined by the data distribution $\rho$ and the augmentation-induced transition matrix $A$. Notably, when $\lambda = 1$, $\pi_{i,j}$ reduces to the \emph{covariance} between the transition probability vectors of $i$ and $j$.

This observation indicates that SCL implicitly performs covariance-based clustering over explicit features, grounded solely in the augmentation distribution and data statistics. As a minimal realization of contrastive learning, it provides a mechanistic explanation for the consistent robustness of CL across diverse downstream tasks.

\subsection{Convergence Target of InfoNCE}

In this section, we analyze the gradient behavior of the InfoNCE loss. Compared to the SCL, InfoNCE not only follows the contrastive learning principle but also implicitly maximizes the mutual information between augmented views. Due to its more complex form, we begin by computing its gradient.

Let $\mathbb{P}_{i,j}$ denote the probability that features $i$ and $j$ are drawn from the same underlying distribution:
\begin{align}
    \mathbb{P}_{i,j} = \frac{\exp\big(sim(z_{i}, z_{j}) / \tau\big)}{\sum_{k \neq j} \exp\big(sim(z_{i}, z_k) / \tau\big) + \exp\big(sim(z_{i}, z_j) / \tau\big)}\,.
\label{eq7}
\end{align}

The gradient of the InfoNCE loss (defined in Eq.~\ref{eq3}) with respect to similarity values is given by:
\begin{align}
    \left\{
    \begin{aligned}
        \frac{\partial \mathcal{L}(G)}{\partial sim(z^+_{i}, z^-_{k})} &= \frac{1}{\tau} \, \mathbb{P}_{i,k}\,, \\
        \frac{\partial \mathcal{L}(G)}{\partial sim(z^+_{i}, z^+_{j})} &= -\frac{1}{\tau}\big(1 - \mathbb{P}_{i,j}\big)\,.
    \end{aligned}
    \right.
\end{align}

Substituting these expressions into the expectation over feature pairs, the gradient becomes:
\begin{align}
    \mathbb{E}[\nabla_\theta \mathcal{L}_{\mathrm{InfoNCE}}] = -\frac{1}{\tau} \left( \pi_p (1 - \mathbb{P}_{i,j}) - \pi_n (n-1) \mathbb{P}_{i,j} \right) \nabla_\theta S\,.
\label{eq10}
\end{align}

To simplify notation, we denote $c_1 = \mathbb{E}[A_i A_j]$ and $c_2 = \mathbb{E}[A_i] \mathbb{E}[A_j]$, and treat them as fixed constants when analyzing the gradient behavior. With this notation, we derive:
\begin{align}
    \mathbb{E}[\nabla^{\mathrm{InfoNCE}}_{i,j;\theta}] &= -\frac{1}{\tau} \left( c_1 (1 - \mathbb{P}_{i,j}) - c_2 (n-1) \mathbb{P}_{i,j} \right) \nabla_\theta S_{i,j} \nonumber \\
    &= -\frac{1}{\tau} \left( c_1 - (c_1 + c_2 (n-1) \mathbb{P}_{i,j}) \right) \nabla_\theta S_{i,j}\,.
 \label{eq11}
\end{align}

The expected gradient direction of InfoNCE evolves throughout training. Under ideal conditions, where the data size tends to infinity, optimizing the InfoNCE loss is equivalent to optimizing this expected gradient. Setting $\mathbb{E}(\nabla_{i,j;\theta}^{InfoNCE}) = 0$ yields $\mathbb{P}_{i,j} = c_1/\left(c_1 + (n-1)\ c_2\right)$ or $\nabla_\theta S_{i,j} = 0$, ignoring the latter. Therefore, for any features $i,j$, InfoNCE has the expectation:
\begin{align}
    \mathbb{P}_{i,j} = \frac{c_1}{\,c_1 + (n-1)\times c_2\,}\,. 
\end{align}
By definition, $0 \leq c_1, c_2 \leq 1$, which implies that $0 \leq \mathbb{P}_{i,j} \leq 1$. To validate its existence, we conduct a synthetic experiment with a known transition probability matrix (setup in  Appendix~\ref{appendix:exp_details}, results in Table~\ref{tab:stm_cmp}). It is worth noting that the empirically estimated probability matrix is not perfectly symmetric due to normalization effects, and thus does not exactly match the theoretical form. Nevertheless, the values closely align with our predictions, confirming the validity of the convergence target. This convergence structure helps explain why InfoNCE typically requires a large number of negative samples or prolonged training. Moreover, its performance may be fundamentally bounded by the discriminative capacity of this target.

\section{Our Method:  Scaled Convergence InfoNCE}

Building on the convergence analysis of InfoNCE, we propose a modified loss that allows explicit control over its convergence target. Since the pairing probabilities $\pi^p_{i,j}$ and $\pi^n_{i,j}$ are determined by the data and augmentation distributions, a weighted form of SCL can be employed to control the convergence behavior. We introduce hyperparameters $\alpha$ and $\beta$ to scale the positive and negative terms in SCL, and define the modified loss as:
\begin{equation}
    \mathcal{L}_{\mathrm{SC\text{-}InfoNCE}} = \mathcal{L}_{\mathrm{InfoNCE}} - \alpha\ \mathrm{sim}(z^+_i, z^+_j)/\tau + \beta\lambda\ \sum^Z \mathrm{sim}(z^+_i, z^-_k)/\tau\ .
\end{equation}

Following the derivations in Equations~\ref{eq10} and~\ref{eq11}, the expected gradient direction becomes:
\begin{align}
\label{eq14}
    \pi_{\mathrm{new}} = -\,c_1(1 - \mathbb{P}_{i,j} + \alpha) + c_2(n-1)(\mathbb{P}_{i,j} + \beta\lambda)\,.
\end{align}
With $\gamma = \beta\lambda$ and $\pi_{\mathrm{new}} = 0$, we obtain:
\begin{equation}
\label{eq15}
    \mathbb{P}_{i,j} = \frac{c_1(1 + \alpha) - c_2\gamma(n - 1)}{c_1 + (n - 1)c_2}\,.
\end{equation}

Next, we discuss the feasible ranges of the hyperparameters \(\alpha\) and \(\gamma\), which control the convergence behavior of our loss. To preserve contrastive properties, we require:
(i) the expected gradient pulls positives together and pushes negatives apart; and
(ii) the gradient for a positive pair is smaller than that for a negative.
From Eq.~\ref{eq14}, these conditions imply:
\begin{equation}
\label{eq:cond-raw}
\begin{aligned}
    -c_1(1 - \mathbb{P}_{i,j} + \alpha) &< c_2(n - 1)(\mathbb{P}_{i,j} + \gamma), \\
    -\frac{1}{\tau}(1 - \mathbb{P}_{i,j} + \alpha)\nabla_\theta S_{i,j} &< \frac{1}{\tau}(\mathbb{P}_{i,k} + \gamma)\nabla_\theta S_{i,k},
\end{aligned}
\end{equation}
where \(k\) is a negative of \(i\). Define \(\delta = 1 + \alpha - \mathbb{P}_{i,j}\), then Eq.~\ref{eq:cond-raw} simplifies to:
\begin{equation}
\label{eq17}
\left\{
\begin{aligned}
    -\frac{c_1}{(n - 1)c_2} \delta &< \mathbb{P}_{i,j} + \gamma, \\
    -\frac{\nabla_\theta S_{i,j}}{\nabla_\theta S_{i,k}} \delta &< \mathbb{P}_{i,k} + \gamma.
\end{aligned}
\right.
\end{equation}

The ratios \(\frac{c_1}{(n-1)c_2}\) and \(\frac{\nabla_\theta S_{i,j}}{\nabla_\theta S_{i,k}}\) are difficult to estimate and may fluctuate during training, making it challenging to determine exact thresholds. As a practical alternative, we adopt a sufficient but non-exhaustive condition by setting \(\delta \ge 0\) and \(\gamma \ge 0\), which ensures convergence in most cases. Under this condition, the convergence target in Eq.~\ref{eq15} becomes:
\begin{equation}
\label{eq:target}
\mathbb{P}_{i,j} = \frac{c_1}{(n - 1)c_2} \delta - \gamma,
\end{equation}
where \(\delta\) serves as a scaling factor and \(\gamma\) as a bias. Note that \(\mathbb{P}_{i,j}\) represents the probability that \(i\) and \(j\) originate from the same distribution, its value must lie in \([0, 1]\). Excessive scaling or bias may result in poor alignment. Regardless, we have obtained a contrastive loss with a scaled convergence target:
\begin{equation}
\mathcal{L}_{\mathrm{SC\text{-}InfoNCE}} = \mathcal{L}_{\mathrm{InfoNCE}} - \frac{1}{\tau} \left( \alpha \  \mathrm{sim}(z^+_i, z^+_j) - \gamma \sum^Z \mathrm{sim}(z^+_i, z^-_k) \right),
\end{equation}
where \(\alpha = \mathbb{P}_{i,j} - 1 + \delta \ge \mathbb{P}_{i,j} - 1\), and \(\gamma \ge 0\).

\section{Experiments}
In this section, we validate the theoretical insights through extensive experiments and provide a comprehensive empirical evaluation of SC-InfoNCE. We first assess its generalization performance on diverse real-world datasets. We then analyze the convergence behavior of different contrastive losses during training. Finally, we conduct ablation studies to examine the impact of the scaling factor on model convergence and stability.

\subsection{Datasets and Experimental Setup}

We evaluate SC‑InfoNCE across four image benchmarks (CIFAR‑10~\cite{cifar}, CIFAR‑100 \cite{cifar}, STL‑10 \cite{stl10}, ImageNet‑100~\cite{imagenet}), four graph benchmarks (COLLAB~\cite{collab}, DD~\cite{tudataset}, NCI1~\cite{nciprotein}, PROTEINS~\cite{nciprotein}), two text benchmarks (STS-B~\cite{stsb}, SICK-R~\cite{sickr}), and a synthetic graph dataset constructed to validate convergence predictions. Baselines include SCL~\cite{ucl}, InfoNCE (as instantiated in SimCLR~\cite{simclr}, GraphCL~\cite{graphcl}, and SimCSE~\cite{simcse}), DCL~\cite{dcl}, DHEL~\cite{kernelana}, and \emph{f}-MICL~\cite{fmicl}.

All methods adopt identical architectures, data augmentation pipelines, optimizers, and batch sizes; the only difference lies in the pretraining objective. ResNet‑50 is used for image tasks, a 3-layer GCN for graph tasks, and BERT-base for text. Each model is pretrained with its respective objective and evaluated via linear probing. Results are averaged over five runs with different random seeds. Full implementation details, including hyperparameter configurations, training schedules, and synthetic dataset construction, are provided in Appendix~\ref{appendix:exp_details}.

\begin{table}[t]
    \centering
    \caption{Performance comparison of different contrastive losses on real-world image, graph, and text datasets. Bold font indicates the best result, and underlined font indicates the second best.}
    \label{tab:true_dataset}
    \resizebox{\textwidth}{!}{%
    \begin{tabular}{lccccccc}
        \toprule
        \multirow{2}*{\textbf{Dataset}} & \multicolumn{5}{c}{\textbf{Baselines}} & \multicolumn{2}{c}{\textbf{Scaled InfoNCE}} \\
        \cmidrule(lr){2-6} \cmidrule(l){7-8}
        & SCL & InfoNCE & DCL & DHEL & \textit{f}-MICL & $\delta$=1.0 & Best \\
        \midrule
        CIFAR-10   & 62.79{\tiny$\pm$0.78} & 90.53{\tiny$\pm$0.38} & 90.69{\tiny$\pm$0.25} & 90.10{\tiny$\pm$0.24} & 90.49{\tiny$\pm$0.27} & \underline{91.28{\tiny$\pm$0.18}} & \textbf{91.49{\tiny$\pm$0.15}} \\
        CIFAR-100  & 35.61{\tiny$\pm$0.52} & 50.90{\tiny$\pm$0.35} & 51.33{\tiny$\pm$0.21} & \underline{51.56{\tiny$\pm$0.25}} & \underline{51.58{\tiny$\pm$0.24}} & 50.71{\tiny$\pm$0.27} & \textbf{51.95{\tiny$\pm$0.19}} \\
        ImageNet-100 & 34.68{\tiny$\pm$0.59} & 74.62{\tiny$\pm$0.29} & 74.22{\tiny$\pm$0.30} & 73.56{\tiny$\pm$0.28} & 74.28{\tiny$\pm$0.28} & \underline{75.38{\tiny$\pm$0.22}} & \textbf{75.62{\tiny$\pm$0.17}} \\
        STL-10     & 51.94{\tiny$\pm$0.64} & 84.07{\tiny$\pm$0.32} & \underline{85.21{\tiny$\pm$0.28}} & 84.99{\tiny$\pm$0.22} & 84.81{\tiny$\pm$0.23} & 85.04{\tiny$\pm$0.26} & \textbf{85.54{\tiny$\pm$0.21}} \\
        \midrule
        COLLAB     & 74.36{\tiny$\pm$1.42} & \underline{75.98{\tiny$\pm$1.11}} & \underline{76.04{\tiny$\pm$1.41}} & 75.24{\tiny$\pm$1.62} & 74.76{\tiny$\pm$1.57} & 75.36{\tiny$\pm$1.68} & \textbf{76.28{\tiny$\pm$1.40}} \\
        DD         & \underline{75.63{\tiny$\pm$4.71}} & 73.92{\tiny$\pm$6.32} & 74.36{\tiny$\pm$3.60} & 73.17{\tiny$\pm$4.93} & 74.10{\tiny$\pm$4.46} & 75.38{\tiny$\pm$4.24} & \textbf{75.89{\tiny$\pm$3.93}} \\
        NCI1       & 74.50{\tiny$\pm$2.13} & 75.38{\tiny$\pm$2.28} & 74.89{\tiny$\pm$2.73} & 74.60{\tiny$\pm$2.41} & \underline{75.62{\tiny$\pm$2.93}} & 75.21{\tiny$\pm$2.43} & \textbf{75.72{\tiny$\pm$2.17}} \\
        PROTEINS   & \underline{71.25{\tiny$\pm$5.27}} & 70.44{\tiny$\pm$3.95} & 70.26{\tiny$\pm$4.16} & 70.97{\tiny$\pm$3.65} & 70.68{\tiny$\pm$3.21} & 70.08{\tiny$\pm$3.97} & \textbf{73.41{\tiny$\pm$2.75}} \\
        \midrule
        STS-B & 12.83{\tiny$\pm$2.15}  & 74.95{\tiny$\pm$1.28} & 73.68{\tiny$\pm$1.52} & 73.92{\tiny$\pm$1.73} & \underline{76.62{\tiny$\pm$1.62}} & 73.39{\tiny$\pm$1.72} & \textbf{77.64{\tiny$\pm$0.89}} \\
        SICK-R & 25.67{\tiny$\pm$3.78}  & 73.87{\tiny$\pm$1.59} & 71.75{\tiny$\pm$1.34} & 72.12{\tiny$\pm$1.48} & \underline{74.28{\tiny$\pm$1.48}} & 72.50{\tiny$\pm$1.67} & \textbf{75.48{\tiny$\pm$1.02}} \\
        \bottomrule
    \end{tabular}
    }
\end{table}

\subsection{Results on Real-world Datasets}

In this section, we evaluate the performance of SC-InfoNCE on real-world datasets. Since SC-InfoNCE replaces the convergence target of InfoNCE (Eq.\ref{eq17}), we first assess its robustness, followed by results under scaling. We report two configurations: one without scaling ($\delta=1, \gamma=0$), and one with tuned $(\delta, \gamma)$ values. Table\ref{tab:true_dataset} summarizes the linear evaluation accuracy on all datasets. Without scaling, SC-InfoNCE performs similarly to standard InfoNCE, with minor variation across datasets, highlighting its robustness. With scaling, it achieves the highest accuracy in all cases.

Among baselines, SCL performs poorly on text and image tasks, and is only competitive on graphs. The improvements of recent variants over InfoNCE are mostly within 1 \%. Based on prior results~\cite{dcl}, these differences may further decrease with  larger batch sizes and longer training schedules. We argue that many contrastive loss variants essentially learn feature representations by aligning different explicit probability transition matrices over feature pairs. When the augmentation distribution is fixed and no additional supervision is introduced, modifying the loss function alone appears to have limited effect on downstream performance.

\begin{figure}[t]
    \centering
    \includegraphics[width=1.0\textwidth]{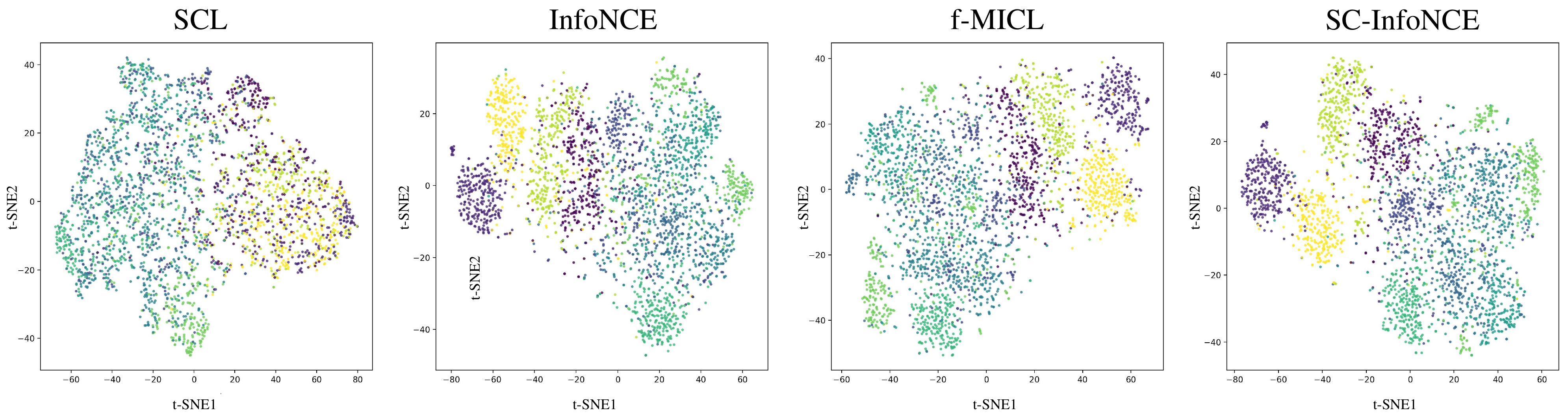}
    \caption{The t-SNE visualizations of representations learned by SCL, InfoNCE, f-MICL, and SC-InfoNCE (ours) on CIFAR-10 show that SC-InfoNCE yields tighter intra-class clusters and clearer inter-class separation.}
    \label{fig_tSNE}
\end{figure}

\begin{figure}[t]
    \centering
    \includegraphics[width=1.0\textwidth]{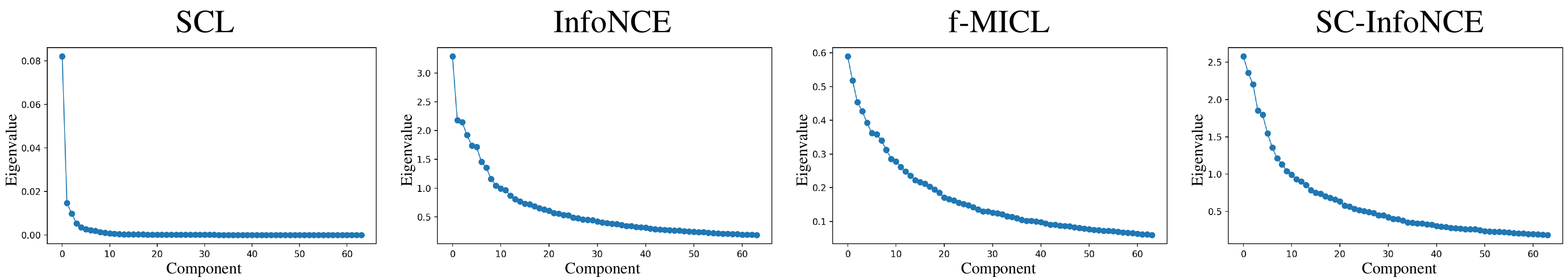}
    \caption{The eigenvalue spectra of the feature covariance matrices for representations learned by SCL, InfoNCE, f-MICL, and SC-InfoNCE (ours) on CIFAR-10. Each curve shows the ordered eigenvalues (in descending order) of the covariance matrix computed from L2-normalized embeddings on the validation set.}
    \label{fig_spectrum}
\end{figure}

\subsection{Representation Analysis}
\label{exp:repa}

In this section, We further examine the learned representations on CIFAR-10 using t-SNE visualizations and feature-spectrum diagnostics, comparing four methods: SCL, InfoNCE, f-MICL, and SC-InfoNCE. As shown in Figure~\ref{fig_tSNE}, SCL produces relatively dispersed clusters, while InfoNCE and f-MICL exhibit noticeable inter-class overlap; in contrast, SC-InfoNCE yields tighter intra-class cohesion and clearer inter-class separation. In the feature-spectrum results of Figure~\ref{fig_spectrum}, SCL concentrates a large portion of variance in the first few principal components, indicating potential collapse; InfoNCE and f-MICL distribute variance more smoothly yet remain limited. By comparison, SC-InfoNCE presents a more balanced spectrum, suggesting a more effective utilization of the representation space. Taken together, these two analyses indicate that, by appropriately scaling the convergence target, SC-InfoNCE not only improves clustering quality but also mitigates representation collapse, leading to overall more robust representations.

\begin{figure}[t]
    \centering
    \includegraphics[width=1.0\textwidth]{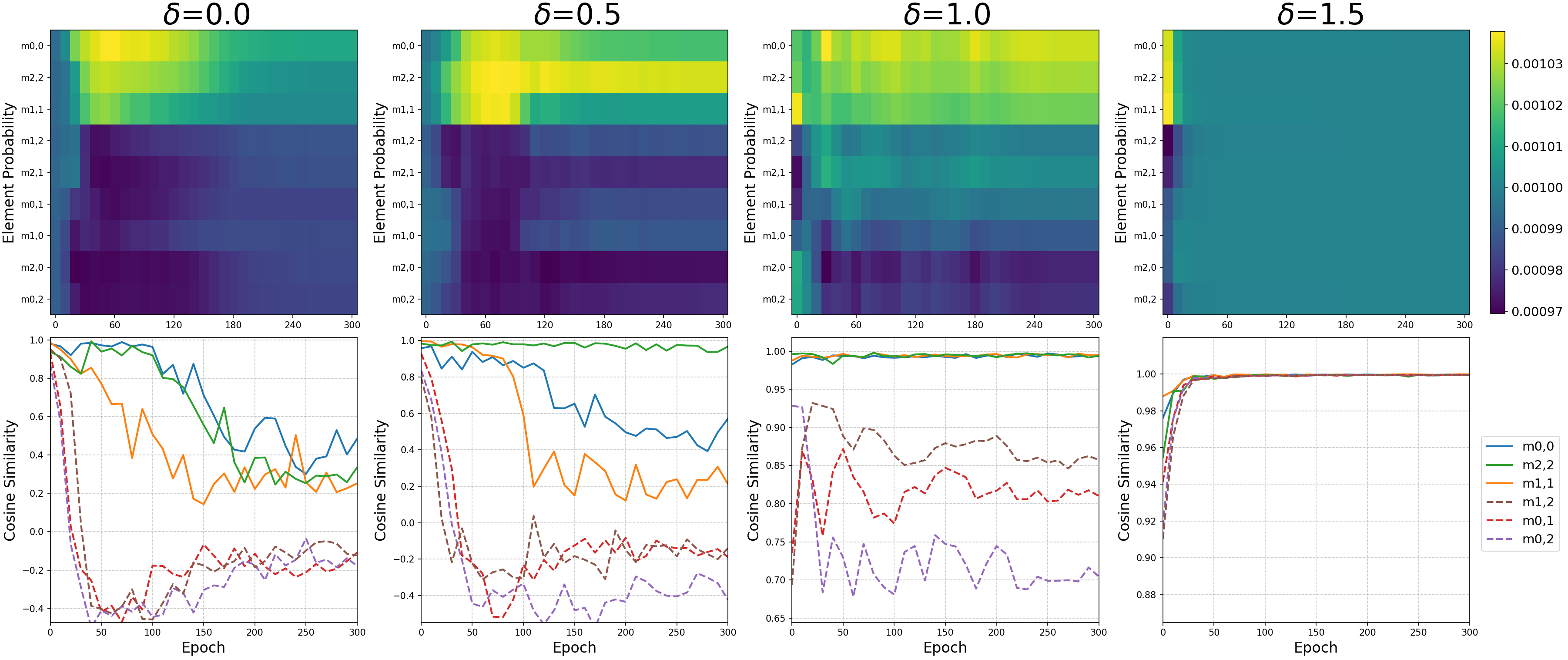}
    \caption{Effect of scaling the convergence target on the convergence trajectory. The vertical ordering of $m_{i,j}$ entries is consistent with their theoretical counterparts $\mathbb{P}_{ij}$.}
    \label{fig2}
\end{figure}

\begin{figure}[t]
    \centering
    \includegraphics[width=1.0\textwidth]{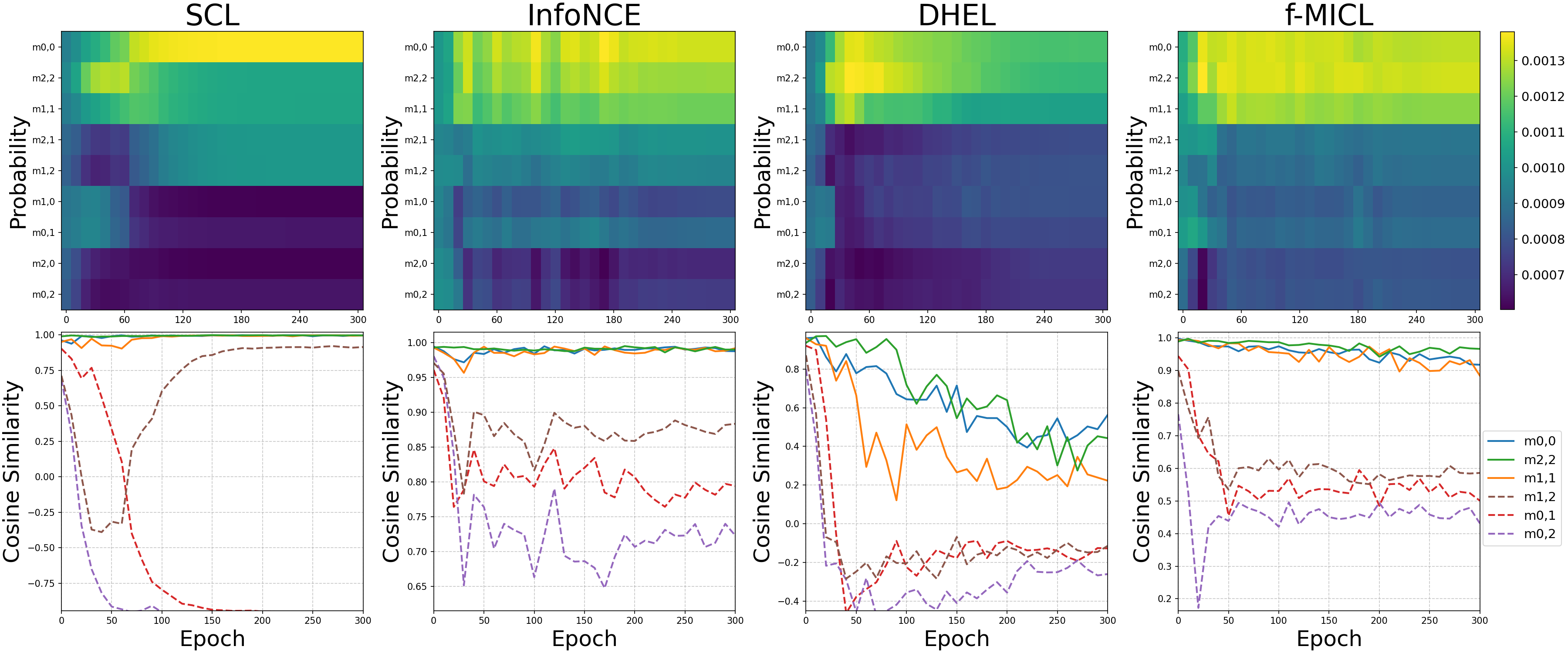}
    \caption{Convergence trajectories for different contrastive losses on the synthetic dataset, visualized using class confusion probabilities and class-level embedding similarities. The vertical ordering of $m_{i,j}$ entries is consistent with their theoretical counterparts $\mathbb{P}_{ij}$.}
    \label{fig1}
\end{figure}

\subsection{Convergence Dynamics Analysis}
\label{exp:cda}
In this section, we first analyze the impact of the scaling factor~$\delta$ on convergence dynamics before comparing different contrastive loss functions. To this end, we conduct ablation studies by evaluating SC-InfoNCE with varying~$\delta$ while fixing $\lambda = 0$. As shown in Figure~\ref{fig2}, reducing the convergence target makes the model more sensitive to fine-grained differences but weakens inter-class separability, whereas increasing the target produces the opposite effect. Excessive scaling beyond a certain threshold, however, leads to feature collapse, where all similarities converge toward one. Notably, when $\delta = 0.5$, SC-InfoNCE successfully separates all six curves in the correct order. These results confirm the positive role of the scaling factor in shaping the final representation structure. Nevertheless, unlike the close match in Table~\ref{tab:stm_cmp}, the estimated probabilities deviate from the modified theoretical values, possibly due to limitations in the estimation method.  

We then compare the convergence dynamics of different contrastive objectives. As illustrated in Figure~\ref{fig1}, we visualize the trajectories using class confusion probabilities and class-level embedding similarities. The vertical ordering of $m_{ij}$ corresponds to the descending order of predicted $\mathbb{P}_{ij}$ values from Table~\ref{tab:stm_cmp}. The heatmap and similarity curves jointly reflect the learned inter-class structure and its alignment with theoretical predictions. Interestingly, by adjusting $\delta$, SC-InfoNCE can approximate the convergence dynamics of other contrastive losses: smaller targets resemble the behavior of DHEL with higher sensitivity to fine-grained differences, while larger targets show closer alignment with InfoNCE, emphasizing inter-class separation. In terms of overall comparison, InfoNCE achieves clear intra-class compactness and inter-class separation. SCL performs poorly due to its coarse sample partitioning, while DHEL exhibits weaker inter-class separability but greater sensitivity to subtle distinctions. F-MICL lies between InfoNCE and DHEL. Overall, despite variations in structural details, all methods except DCL successfully distinguish samples from different classes.  

Finally, we further examine the effect of common hyperparameters such as batch size and temperature on InfoNCE convergence to validate conclusions from prior work~\cite{ucl,simclr}. Detailed results are reported in Appendix~\ref{apd:exp}.

\section{Related Work}

Contrastive learning (CL) originates from Siamese networks \cite{1992,2005,2006} and the $(N{+}1)$‑tuple formulation of \cite{2016}, and rose to prominence when InfoNCE in Contrastive Predictive Coding (CPC) \cite{infonce} framed CL as mutual‑information maximization. Vision systems such as SimCLR \cite{simclr} and MoCo \cite{moco,moco2} combined large‑batch training, momentum encoders, and rich data augmentations, while hard‑negative mining further boosted performance \cite{negcl,negcl2,cmc,ref11,dcl,ref12}. The paradigm soon spread to natural language \cite{cert,consert,simcse,simcse++}, graphs \cite{graphcl,joao,ref13,ref9,ref14}, and multimodal tasks \cite{ref15,signclip}, inspiring augmentation‑free variants \cite{byol,ref10} and label‑aware objectives \cite{scl,ref19}. Complementary studies probe how the \emph{number of negatives} \cite{dcl,ref1}, the \emph{temperature} hyper‑parameter \cite{ref3,ucl}, and anti‑collapse techniques \cite{collapse} affect training efficacy; however, these works focus on empirical heuristics and leave open a principled account of how InfoNCE organizes features and induces cluster structure, which is the central question this paper seeks to address.

\section{Extension and Application}

From the augmentation-kernel perspective adopted in this work, the convergence behavior of contrastive learning can be effectively characterized by a TPM or its suitable approximation. In practice, as feature spaces are high-dimensional and augmentation mechanisms complex, constructing a full TPM is often infeasible and unnecessary. For a specific downstream task, a finite sub-TPM can be derived from task priors and class-discrimination requirements, focusing on key patterns and augmentation channels that strongly influence prediction. A typical procedure includes:
(i) defining task scope and identifying task-relevant observable patterns;
(ii) designing limited augmentations around these patterns and estimating empirical transition frequencies by repeated sampling;
(iii) applying temperature scaling, smoothing, and sparsity control (e.g., k-nearest neighbors), followed by row normalization; and
(iv) using the sub-TPM to guide batch construction, positive/negative sampling, and invariance regularization, while reweighting to mitigate class imbalance. This approach offers interpretable control over the effects of difficult or imbalanced samples and stabilizes representation learning.

Beyond a specific contrastive-loss form, the framework can be abstracted to a broader class of learning problems. When a task admits an observable and approximately stationary augmentation mechanism, and its label semantics align with the induced invariance, a TPM-induced structural prior can be defined in the explicit feature space. This structure captures augmentation-based equivalence relations and inter-class separability, providing a unified theoretical lens for analyzing invariance and discriminability. In this sense, TPMs extend beyond contrastive learning to serve as general representational priors that encode the statistical structure of data augmentations. Different tasks may selectively incorporate such priors—at either the objective or data-organization level—depending on their feature observability and invariance assumptions. This conceptual generalization suggests new directions for exploring the theoretical role and applicability of TPMs across learning paradigms.

\section{Conclusion}
In this paper, we introduce an explicit feature space to characterize and predict the convergence behavior of the InfoNCE loss. Theoretical analysis and empirical evidence show that InfoNCE implicitly aligns the pairwise probability matrix over features with a transition matrix determined by the dataset and augmentation distribution, thereby enabling unsupervised instance-level discrimination. Building on this insight, we propose SC-InfoNCE, a contrastive loss with a tunable convergence target, and validate its effectiveness on both synthetic and real-world datasets. While SC-InfoNCE improves controllability and performance, its current formulation applies global target scaling and is constrained by asymmetric probability estimation and normalization, which hinder full convergence to the designed target. Future work may explore symmetrized probability formulations or data-driven adaptive scaling strategies to address these limitations.

\section*{Acknowledgements}
This work was supported by the Hunan Province Key R\&D Project (Grant No. 2024JK2004) and the Graduate Student Research and Innovation Project of Xiangtan University (Grant No. XDCX2024Y278).

\section{Appendix}

\subsection{Non-Independent Positive and Negative Sampling}
\label{apd:the}

In practice, the sampling of positive and negative pairs in frameworks such as SimCLR and GraphCL is not strictly independent. Specifically, when a feature appears in both the positive and negative sets for the same anchor, their contributions to the gradient cancel out. As a result, some samples yield no effective learning signal. This behavior implies that the computation of $\pi$ in Eq. \ref{eq6} does not fully capture the actual dynamics during training. A more precise formulation of $\pi_{i,j}$ that accounts for this overlap is given by:
\begin{footnotesize}
\begin{align}
    \pi_{i,j} &= \sum_{m=1,\,m \neq i,j}^{|F|}\;\sum_{n=1}^{|F|}(P_n A_{n,m}) \sum_{k=1}^{|F|}(P_k A_{k,i} A_{k,j}) - \sum_{l=1}^{|F|}(P_l A_{l,j}) \sum_{o=1,\,o \neq j}^{|F|} \sum_{k=1}^{|F|}(P_k A_{k,i} A_{k,o}) \nonumber \\
    &= \sum_{k=1}^{|F|}\sum_{n=1}^{|F|} P_k A_{k,i} P_n \Big( \sum_{m=1,\,m \neq i,j}^{|F|} A_{n,m} A_{k,j} - \sum_{m=1,\,m \neq j}^{|F|} A_{n,j} A_{k,m} \Big) \nonumber \\
    &= \sum_{n=1}^{|F|} P_n \sum_{k=1}^{|F|} P_k\, C_{ij}^{\,nk},
\end{align}
\end{footnotesize}
where $C_{ij}^{nk}$ is a constant that does not depend on the current prediction distribution $P_c$.

Importantly, this correction does not affect the main conclusions presented in the paper. When the dataset and augmentation distribution are sufficiently diverse, the effect of sample overlap and gradient cancellation becomes negligible. For this reason, and to maintain analytic clarity, we adopt the assumption of independent sampling for positives and negatives in the main analysis.

\subsection{Error Analysis}
In this subsection we provide an error analysis for the TPM. Following Section \ref{sec3.1}, let the sampling probability vector be \(P\in\Delta^{m-1}\) (i.e., \(\|P\|_1=1\) and \(P>0\)), where the total number of features is \(m=|F\cup F_+|\). Let the transition probability matrix be \(A\in[0,1]^{m\times m}\). For any feature indices \(i,j\), define
\begin{equation}
c_1:=\mathbb{E}[A_iA_j]=P^\top(A_i\odot A_j),\,
c_2:=\mathbb{E}[A_i]\mathbb{E}[A_j]=s_is_j,\,
s_k:=P^\top A_k,
\end{equation}
and set
\begin{equation}
\mathbb{P}_{i,j}:=\frac{c_1}{D},\,
D:=c_1+(n-1)c_2,\,
\delta_i:=\hat c_i-c_i,\, \epsilon_p:=\hat P-P.
\end{equation}

Performing a first-order expansion of \(\hat c_1=\hat P^\top(\hat A_i\odot \hat A_j)\) (neglecting second-order terms \(\epsilon_p\odot \delta A\) and \(\delta A_i\odot\delta A_j\)) yields
\begin{equation}
\delta_1
=\epsilon_p^\top(A_i\odot A_j)
+P^\top(\delta A_i\odot A_j)
+P^\top(A_i\odot \delta A_j).
\end{equation}
Since \(0\le A\le 1\) entrywise and \(\|P\|_1=1\), we obtain
\begin{equation}
|\delta_1|\le \|\epsilon_p\|_1+s_j\|\delta A_i\|_\infty+s_i\|\delta A_j\|_\infty.
\end{equation}

Similarly, from \(\delta_2=s_i\,\delta s_j+\delta s_i\,s_j+\delta s_i\,\delta s_j\) and the bound
\(|\delta s_k|=|\epsilon_p^\top A_k+P^\top\delta A_k|\le \|\epsilon_p\|_1+\|\delta A_k\|_\infty\),
we obtain the first-order upper bound
\begin{equation}
|\delta_2|\le (s_i+s_j)\|\epsilon_p\|_1+s_j\|\delta A_i\|_\infty+s_i\|\delta A_j\|_\infty.
\end{equation}

For uniform control over all pairs \((i,j)\), let
\[
s_{\min}:=\min_k s_k,\qquad
\eta_{\max}:=\max_k \|\delta A_k\|_\infty.
\]
Then the bounds simplify to
\begin{equation}
|\delta_1|\le \|\epsilon_p\|_1+2\eta_{\max},\qquad
|\delta_2|\le 2\|\epsilon_p\|_1+2\eta_{\max}.
\end{equation}

Let \(\delta D:=\delta_1+(n-1)\delta_2\). Since
\begin{equation}
\hat{\mathbb{P}}_{i,j}-\mathbb{P}_{i,j}
=\frac{\delta_1D-c_1\delta D}{D(D+\delta D)},
\end{equation}
and \(c_1\le D\), whenever \(D>|\delta D|\) we have
\begin{equation}
\bigl|\hat{\mathbb{P}}_{i,j}-\mathbb{P}_{i,j}\bigr|
\le \frac{2|\delta_1|+(n-1)|\delta_2|}{D-|\delta D|}.
\end{equation}
On the other hand,
\begin{equation}
D=c_1+(n-1)c_2\ge (n-1)s_is_j\ge (n-1)s_{\min}^2,
\end{equation}
and
\begin{equation}
|\delta D|\le |\delta_1|+(n-1)|\delta_2|
\le 2n\|\epsilon_p\|_1+2n\eta_{\max}.
\end{equation}
Hence a sufficient condition for a positive denominator margin is
\begin{equation}
D-|\delta D|\ \ge\ \tfrac12 (n-1)s_{\min}^2
\ \Longleftarrow\
2n\|\epsilon_p\|_1+2n\eta_{\max}\ \le\ \tfrac12 (n-1)s_{\min}^2 .
\end{equation}
Under this condition, the upper bound specializes to
\begin{footnotesize}
\begin{align}
\max_{i,j}\bigl|\hat{\mathbb{P}}_{i,j}-\mathbb{P}_{i,j}\bigr|\le\
\frac{2n\|\epsilon_p\|_1+(2n+2)\eta_{\max}}{\tfrac12 (n-1)s_{\min}^2}
=\underbrace{\frac{4n}{n-1}}_{=:C_1}\frac{\|\epsilon_p\|_1}{s_{\min}^2}
+\underbrace{\frac{4n+4}{\,n-1\,}}_{=:C_2}\frac{\eta_{\max}}{s_{\min}^2}.
\end{align}
\end{footnotesize}

For the empirical distribution \(\hat P\), we invoke the standard \(\ell_1\) concentration inequality for discrete distributions. For any confidence level \(\delta\in(0,1)\),
\begin{equation}
\Pr\!\Bigl(\|\hat P-P\|_1\le \varepsilon_p(n,\delta)\Bigr)\ \ge\ 1-\delta,\,
\varepsilon_p(n,\delta):=\sqrt{\frac{2\bigl(m\ln 2+\ln(1/\delta)\bigr)}{n}}.
\end{equation}
Substituting \(\|\epsilon_p\|_1\) by \(\varepsilon_p(n,\delta)\) and requiring the right-hand side to be at most a prescribed accuracy \(\varepsilon>0\) yield, with probability at least \(1-\delta\), the sample-complexity and model-error conditions
\begin{equation}
n\ \ge\ \frac{8\,C_1^{2}}{\varepsilon^{2}s_{\min}^{4}}\,
\Bigl(m\ln 2+\ln\tfrac{1}{\delta}\Bigr),
\qquad
\eta_{\max}\ \le\ \frac{\varepsilon\,s_{\min}^{2}}{2\,C_2}.
\end{equation}

For a simpler, \(n\)-free presentation, one may adopt constant upper bounds. In the common regime \(n\ge 3\), take \(C_1=6\) and \(C_2=8\), which gives
\begin{equation}
n\ \ge\ \frac{288}{\varepsilon^{2}s_{\min}^{4}}\,
\Bigl(m\ln 2+\ln\tfrac{1}{\delta}\Bigr),
\qquad
\eta_{\max}\ \le\ \frac{\varepsilon\,s_{\min}^{2}}{16},
\end{equation}
so that
\begin{equation}
\max_{i,j}\bigl|\hat{\mathbb{P}}_{i,j}-\mathbb{P}_{i,j}\bigr|\ \le\ \varepsilon
\end{equation}
holds with probability at least \(1-\delta\). The foregoing conclusions are based on a first-order approximation and neglect higher-order perturbation terms.

\subsection{Experimental Details}
\label{appendix:exp_details}
We evaluate SC‑InfoNCE across three contrastive pretraining domains: image, graph, and text. These domains differ substantially in input modality, data scale, and structural characteristics. A summary of dataset statistics is provided in Table~\ref{tab:datasets}. All models are implemented in PyTorch and trained on dual NVIDIA A100 GPUs (80~GB). Full code and configuration files are available at \url{https://anonymous.4open.science/r/SC-InfoNCE-95E3}.

\paragraph{Image Classification} 
We evaluate our method on four standard image classification benchmarks: CIFAR-10, CIFAR-100, STL-10, and ImageNet-100. All experiments use a ResNet-50~\cite{resnet} backbone with a batch size of 256. Pretraining is conducted for 200 epochs on all datasets except STL-10, which is pretrained for 100 epochs. Each pretraining phase is followed by 200 epochs of linear evaluation. The base learning rate is set to 4, except for ImageNet-100 where it is 1, using a \textit{warmup-anneal} schedule. For fine-tuning, the initial learning rates are 0.1, 0.01, 0.01, and 1.6 for CIFAR-10, CIFAR-100, STL-10, and ImageNet-100, respectively, using linear decay. The temperature parameter is set to 0.5 for CIFAR-10 and 0.15 for the other datasets. Data augmentation and split strategies follow those of SimCLR. For the SC-InfoNCE loss, we select $\delta$ from the candidate set $\{-0.1, 0.0, 0.2, 0.5, 1.0, 2.0\}$ and $\gamma$ from $\{-0.2, -0.1, 0.0, 0.1, 0.2\}$, where $\gamma$ is scaled by the number of negatives.

\begin{table}[!htbp]
\centering
\small
\caption{Overview of datasets used in image, graph, and text experiments.}
\label{tab:datasets}
\begin{tabular}{p{1.4cm} p{4.0cm} p{6.4cm}}
\toprule
\textbf{Domain} & \textbf{Dataset} & \textbf{Summary} \\
\midrule
\multirow{3}{*}{Image} 
  & CIFAR-10/100~\cite{cifar} & 32$\times$32 natural images; 60k samples across 10 or 100 classes \\
  & STL-10~\cite{stl10} & 96$\times$96 images; 5k labeled and 100k unlabeled samples; 10-class classification \\
  & ImageNet-100~\cite{imagenet} & First 100 classes from ImageNet (by class index); 67,198 images \\
\midrule
\multirow{4}{*}{Graph}
  & COLLAB~\cite{collab} & Scientific collaboration networks; 5k graphs over 3 classes \\
  & DD~\cite{tudataset} & Protein structure graphs; 1.2k samples, 2-class classification \\
  & NCI1~\cite{nciprotein} & Chemical compound graphs; 4.1k graphs, binary labels \\
  & PROTEINS~\cite{nciprotein} & Protein interaction graphs; 1.1k graphs, 2 classes \\
\midrule
\multirow{3}{*}{Text}
  & STS-B~\cite{stsb} & 8.6k sentence pairs with human-annotated similarity scores (1–5) \\
  & SICK-R~\cite{sickr} & 9.8k sentence pairs with relatedness scores and NLI labels \\
  % & Wiki1M~\cite{simcse} & 1M sentences from Wikipedia for contrastive pretraining \\
\bottomrule
\end{tabular}
\end{table}

\paragraph{Graph and Text Tasks}
For graph classification, we consider four datasets: COLLAB, DD, NCI1, and Proteins. We adopt a 3-layer GCN encoder followed by a 2-layer linear classifier, with a batch size of 128. Models are pretrained for 200 epochs and fine-tuned for 100 epochs, using a fixed learning rate of 0.001 and a temperature of 0.5. Data augmentations and train/test splits follow the GraphCL protocol. The SC-InfoNCE hyperparameter settings match those in the image experiments.

For the text task, we follow the SimCSE setup. Contrastive pretraining is performed on 1M sentences sampled from English Wikipedia, followed by evaluation on STS-B and SICK-R. We use a BERT-base~\cite{bert} encoder with a batch size of 64, pretraining for 1 epoch and fine-tuning for 4 epochs. The learning rate is fixed at $3 \times 10^{-5}$, and the temperature is set to 0.05. Other details, including augmentations and data splits, follow SimCSE. The $(\delta, \gamma)$ values for SC-InfoNCE are consistent with those in prior experiments.

\begin{figure}[t]
    \centering
    \includegraphics[width=1.0\textwidth]{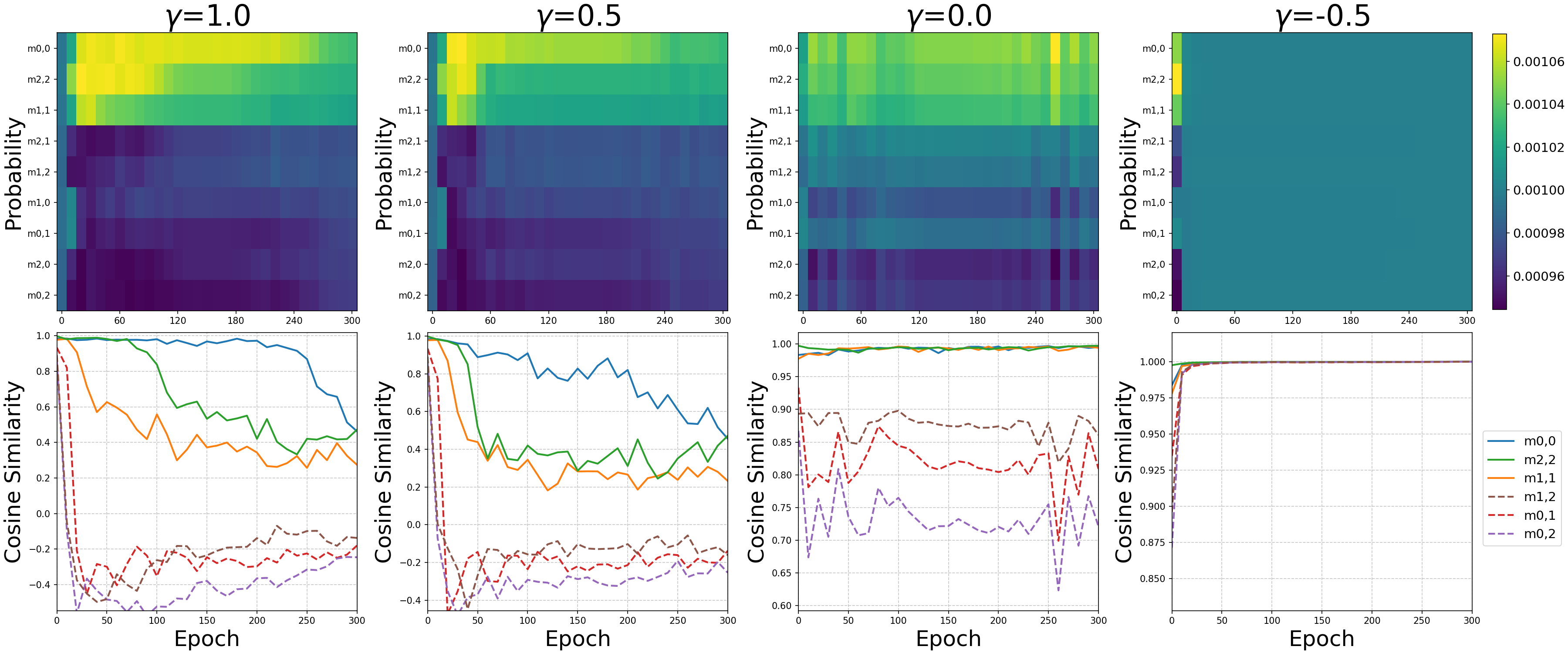}
    \caption{Convergence comparison under different values of $\gamma$ (normalized by batch size) with fixed $\delta=1$ on a synthetic dataset.}
    \label{fig:gamma}
\end{figure}

\paragraph{Synthetic Graph Classification}
To validate our theoretical convergence analysis, we construct a synthetic graph classification benchmark with a known transition probability matrix. We generate three graph classes (labels 0, 1, 2), each with an average of 20 nodes and 40 edges. Node features are one-hot vectors; edge features are omitted. In each instance, 20\% of the edges are randomly rewired to introduce noise. A total of 4000 graphs are generated via uniform sampling and split evenly into training and test sets. We simulate class transitions on augmented views using the ground-truth transition matrix, followed by noisy perturbation. We employ a 2-layer GCN encoder with a linear projection head, a feature dimension of 16, a batch size of 1000, and train for up to 300 epochs using the Adam optimizer with a fixed learning rate of 0.001. The temperature is set to 1. Test performance is monitored to analyze convergence trajectories.

\begin{figure}[!htbp]
    \centering
    \includegraphics[width=1.0\textwidth]{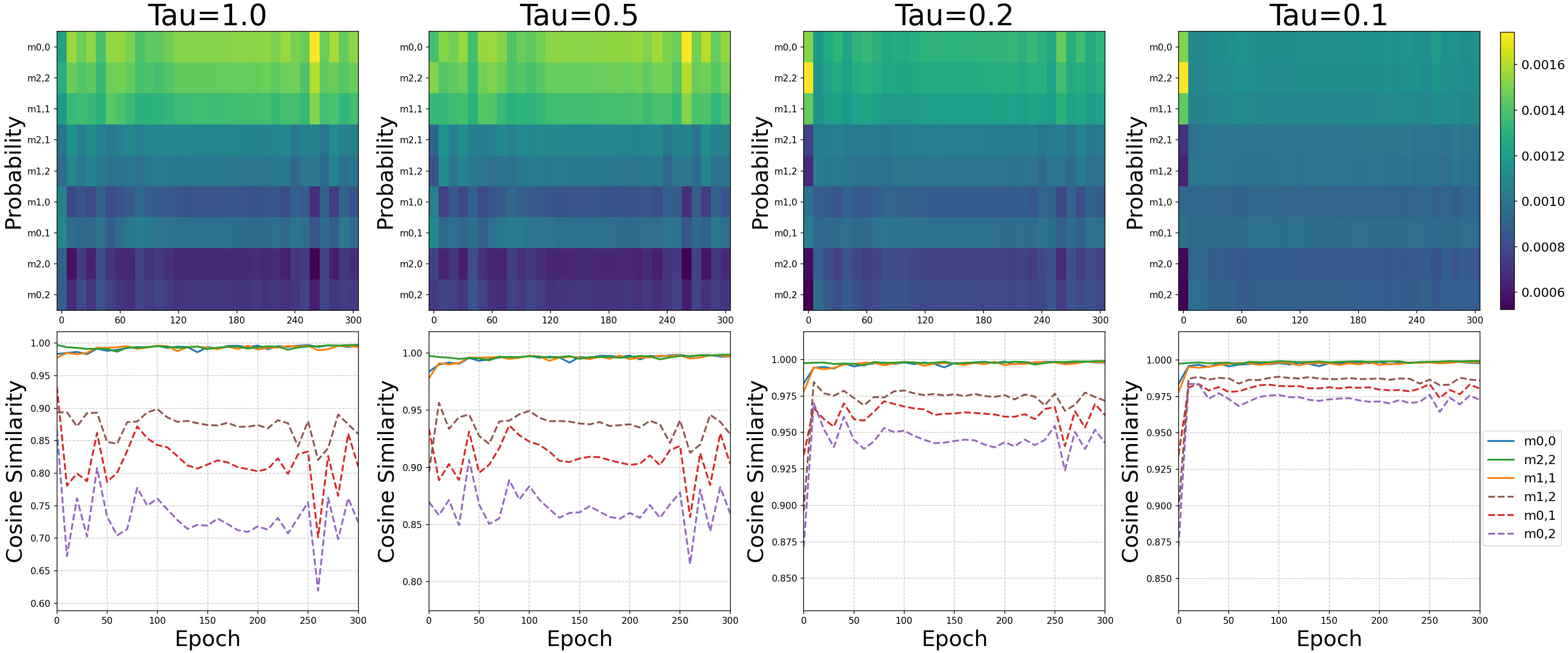}
    \caption{Convergence trajectories of different contrastive losses under different temperatures on a synthetic dataset.}
    \label{fig:tau}
\end{figure}
\begin{figure}[!htbp]
    \centering
    \includegraphics[width=1.0\textwidth]{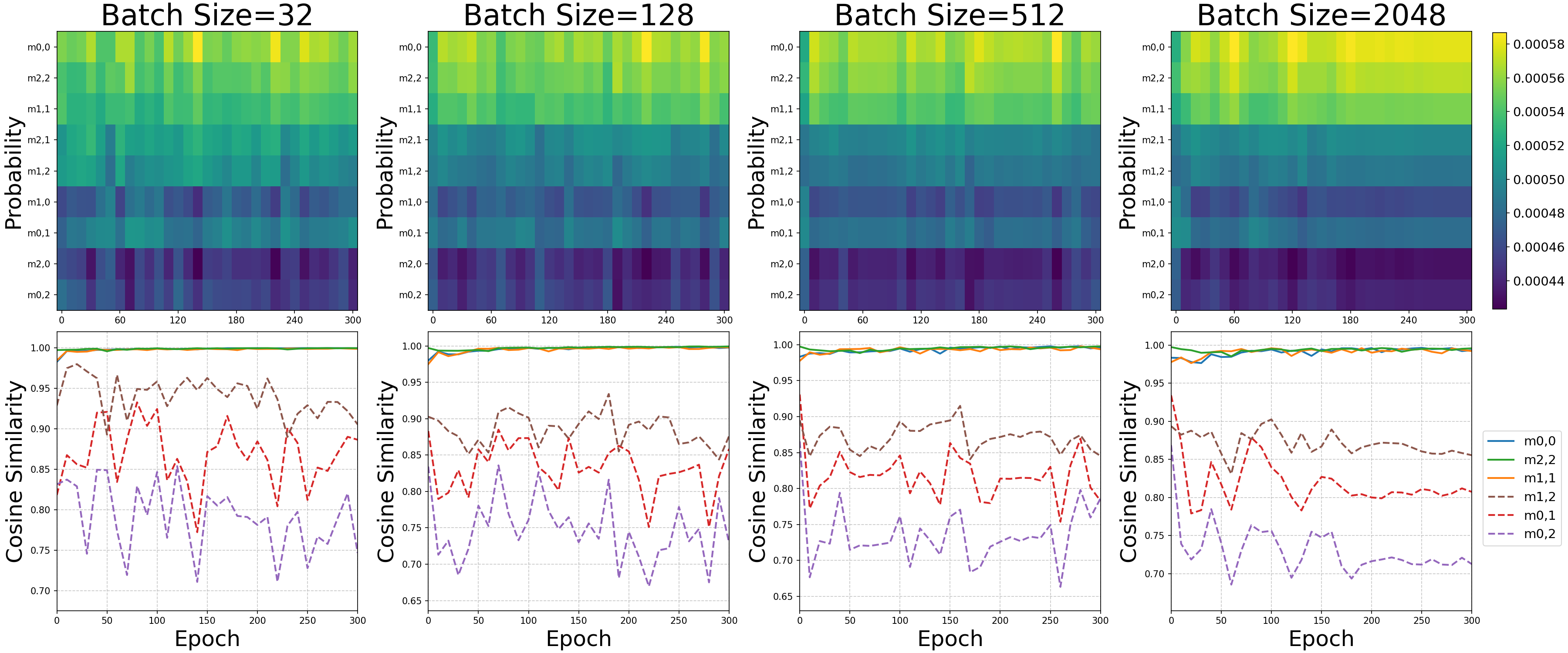}
    \caption{Convergence trajectories of different contrastive losses under different batch sizes on a synthetic dataset.}
    \label{fig:bz}
\end{figure}

\subsection{Additional Experimental Results}
\label{apd:exp}

The performance of contrastive learning has been shown to be highly sensitive to hyperparameters such as temperature and batch size. In this section, we conduct a systematic empirical analysis to further investigate how these hyperparameters influence convergence behavior.

\paragraph{Temperature}  
According to our theory, temperature does not affect the final convergence target, but it is essential for computing pairwise probabilities. A lower temperature makes $\mathbb{P}_{i,j}$ more sensitive to marginal changes in similarity, meaning that a smaller variation is sufficient to reach the target. Figure~\ref{fig:tau} illustrates how the heatmaps and similarity curves evolve as temperature decreases. Empirically, we observe that increasing the temperature systematically decreases $\mathbb{P}_{i,j}$, while the similarity between embeddings from different classes gradually increases.

\begin{figure}[t]
    \centering
    \includegraphics[width=1.0\textwidth]{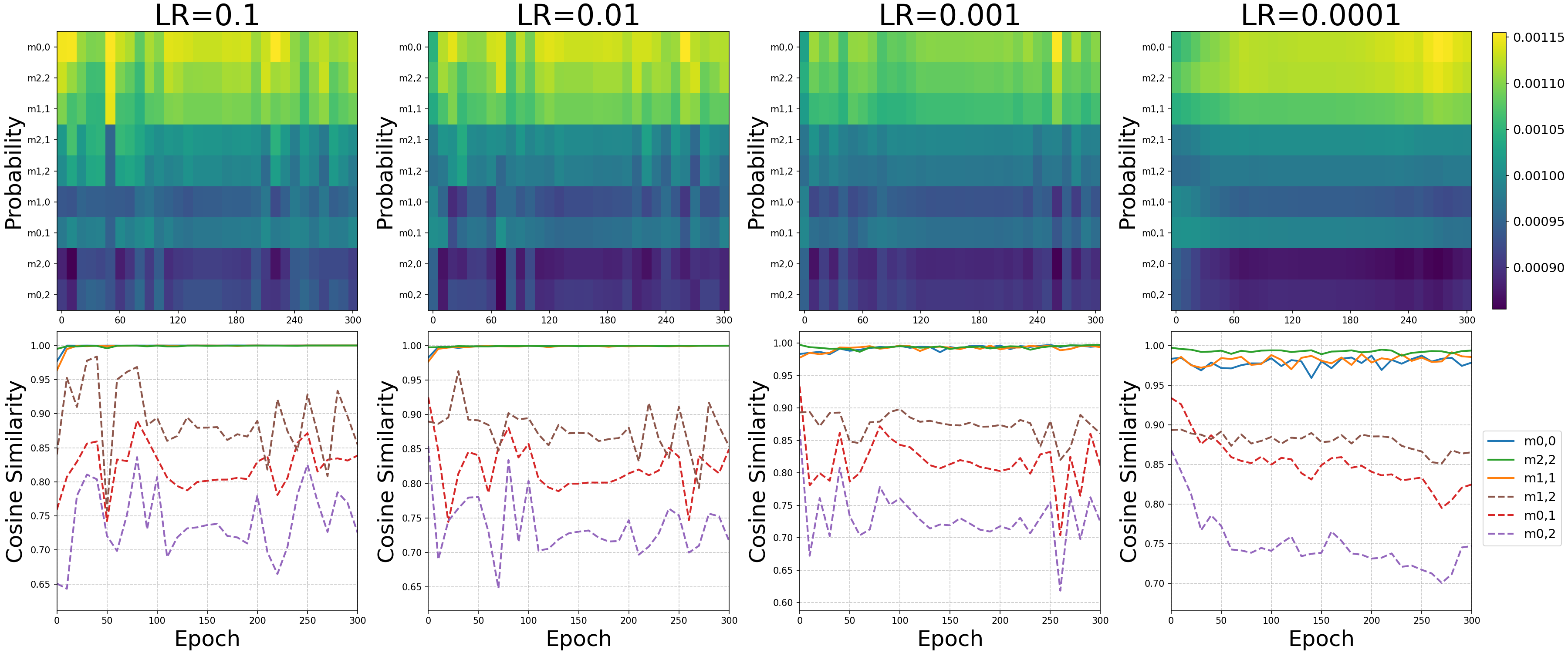}
    \caption{Convergence trajectories of contrastive objectives under different learning rates on a synthetic dataset.}
    \label{fig:lr}
\end{figure}

\paragraph{Batch Size}  
Theoretically, increasing the batch size does not alter the target distribution, but enables a more stable approximation of the expected probabilities. Starting from a batch size of 32, we increase the number of negatives by a factor of 4 at each step. The results, shown in Figure~\ref{fig:bz}, indicate that while the similarity range across classes remains approximately constant, larger batch sizes yield smoother similarity profiles and improved inter-class discrimination. These observations are consistent with our theoretical expectations.

\paragraph{Learning Rate}  
The learning rate is a key hyperparameter that controls the step size in gradient descent. We evaluate its effect in Figure~\ref{fig:lr}. Lower learning rates result in slower decreases in similarity, but the convergence trajectory exhibits greater stability.

\paragraph{Hyperparameter $\gamma$}  
In the main text, we analyzed the effect of varying $\delta$ under the setting $\gamma = 0$. Here, we further extend our analysis by examining the effect of $\gamma$ under $\delta = 1$ (with $\gamma$ normalized as $\gamma / (n - 1)$). As shown in Figure~\ref{fig:gamma}, the convergence curves under both settings exhibit strong agreement, corroborating our theoretical predictions.

% %%       \cite{<label>} ==> [1]

% %%
% Example citation, See \cite{lamport94}.

\bibliographystyle{elsarticle-num} 
\bibliography{cite}

@String{Computing = "Computing" }

@String{Computer = "{IEEE} Computer" }

@String{Springer = "Springer-Verlag" }

@inproceedings{simclr,
  title={A simple framework for contrastive learning of visual representations},
  author={Chen, Ting and Kornblith, Simon and Norouzi, Mohammad and Hinton, Geoffrey},
  booktitle={International conference on machine learning},
  pages={1597--1607},
  year={2020},
  organization={PmLR}
}

@inproceedings{moco,
  title={Momentum contrast for unsupervised visual representation learning},
  author={He, Kaiming and Fan, Haoqi and Wu, Yuxin and Xie, Saining and Girshick, Ross},
  booktitle={Proceedings of the IEEE/CVF conference on computer vision and pattern recognition},
  pages={9729--9738},
  year={2020}
}

@inproceedings{fastmoco,
  title={Fast-MoCo: Boost momentum-based contrastive learning with combinatorial patches},
  author={Ci, Yuanzheng and Lin, Chen and Bai, Lei and Ouyang, Wanli},
  booktitle={European Conference on Computer Vision},
  pages={290--306},
  year={2022},
  organization={Springer}
}

@inproceedings{dcl,
  title={Decoupled contrastive learning},
  author={Yeh, Chun-Hsiao and Hong, Cheng-Yao and Hsu, Yen-Chi and Liu, Tyng-Luh and Chen, Yubei and LeCun, Yann},
  booktitle={European conference on computer vision},
  pages={668--684},
  year={2022},
  organization={Springer}
}

@article{fmicl,
  title={$ f $-MICL: Understanding and Generalizing InfoNCE-based Contrastive Learning},
  author={Lu, Yiwei and Zhang, Guojun and Sun, Sun and Guo, Hongyu and Yu, Yaoliang},
  journal={arXiv preprint arXiv:2402.10150},
  year={2024}
}

@inproceedings{Spectral,
  title={Contrastive Learning is Spectral Clustering on Similarity Graph},
  author={Tan, Zhiquan and Zhang, Yifan and Yang, Jingqin and Yuan, Yang},
  booktitle={ICLR},
  year={2024}
}

@article{kernelana,
  title={Bridging mini-batch and asymptotic analysis in contrastive learning: from InfoNCE to kernel-based losses},
  author={Koromilas, Panagiotis and Bouritsas, Giorgos and Giannakopoulos, Theodoros and Nicolaou, Mihalis and Panagakis, Yannis},
  journal={arXiv preprint arXiv:2405.18045},
  year={2024}
}

@inproceedings{basetheo,
  title={A theoretical analysis of contrastive unsupervised representation learning},
  author={Saunshi, Nikunj and Plevrakis, Orestis and Arora, Sanjeev and Khodak, Mikhail and Khandeparkar, Hrishikesh},
  booktitle={International Conference on Machine Learning},
  pages={5628--5637},
  year={2019},
  organization={PMLR}
}

@inproceedings{ucl,
  title={Understanding the behaviour of contrastive loss},
  author={Wang, Feng and Liu, Huaping},
  booktitle={Proceedings of the IEEE/CVF conference on computer vision and pattern recognition},
  pages={2495--2504},
  year={2021}
}

@article{moco2,
  title={Improved baselines with momentum contrastive learning},
  author={Chen, Xinlei and Fan, Haoqi and Girshick, Ross and He, Kaiming},
  journal={arXiv preprint arXiv:2003.04297},
  year={2020}
}

@article{infonce,
  title={Representation learning with contrastive predictive coding},
  author={Oord, Aaron van den and Li, Yazhe and Vinyals, Oriol},
  journal={arXiv preprint arXiv:1807.03748},
  year={2018}
}

@article{surveygcl,
  title={Towards graph contrastive learning: A survey and beyond},
  author={Ju, Wei and Wang, Yifan and Qin, Yifang and Mao, Zhengyang and Xiao, Zhiping and Luo, Junyu and Yang, Junwei and Gu, Yiyang and Wang, Dongjie and Long, Qingqing and others},
  journal={arXiv preprint arXiv:2405.11868},
  year={2024}
}

@article{byol,
  title={Bootstrap your own latent-a new approach to self-supervised learning},
  author={Grill, Jean-Bastien and Strub, Florian and Altch{\'e}, Florent and Tallec, Corentin and Richemond, Pierre and Buchatskaya, Elena and Doersch, Carl and Avila Pires, Bernardo and Guo, Zhaohan and Gheshlaghi Azar, Mohammad and others},
  journal={Advances in neural information processing systems},
  volume={33},
  pages={21271--21284},
  year={2020}
}

@inproceedings{llmc2024,
  title={Large Language Models can Contrastively Refine their Generation for Better Sentence Representation Learning},
  author={Wang, Huiming and Li, Zhaodonghui and Cheng, Liying and Bing, Lidong and others},
  booktitle={NAACL-HLT},
  year={2024}
}

@inproceedings{simcse,
  title={SimCSE: Simple Contrastive Learning of Sentence Embeddings},
  author={Gao, T and Yao, X and Chen, Danqi},
  booktitle={EMNLP 2021-2021 Conference on Empirical Methods in Natural Language Processing, Proceedings},
  year={2021}
}

@article{infobtn,
  title={The information bottleneck method},
  author={Tishby, Naftali and Pereira, Fernando C and Bialek, William},
  journal={arXiv preprint physics/0004057},
  year={2000}
}

@article{ddpm,
  title={Denoising diffusion probabilistic models},
  author={Ho, Jonathan and Jain, Ajay and Abbeel, Pieter},
  journal={Advances in neural information processing systems},
  volume={33},
  pages={6840--6851},
  year={2020}
}

@inproceedings{ldm,
  title={High-resolution image synthesis with latent diffusion models},
  author={Rombach, Robin and Blattmann, Andreas and Lorenz, Dominik and Esser, Patrick and Ommer, Bj{\"o}rn},
  booktitle={Proceedings of the IEEE/CVF conference on computer vision and pattern recognition},
  pages={10684--10695},
  year={2022}
}

@article{graphcl,
  title={Graph contrastive learning with augmentations},
  author={You, Yuning and Chen, Tianlong and Sui, Yongduo and Chen, Ting and Wang, Zhangyang and Shen, Yang},
  journal={Advances in neural information processing systems},
  volume={33},
  pages={5812--5823},
  year={2020}
}

@article{cifar,
  title={Learning Multiple Layers of Features from Tiny Images},
  author={Krizhevsky, A},
  journal={Master's thesis, University of Tront},
  year={2009}
}

@inproceedings{stl10,
  title={An analysis of single-layer networks in unsupervised feature learning},
  author={Coates, Adam and Ng, Andrew and Lee, Honglak},
  booktitle={Proceedings of the fourteenth international conference on artificial intelligence and statistics},
  pages={215--223},
  year={2011},
  organization={JMLR Workshop and Conference Proceedings}
}

@article{imagenet,
  title={Imagenet large scale visual recognition challenge},
  author={Russakovsky, Olga and Deng, Jia and Su, Hao and Krause, Jonathan and Satheesh, Sanjeev and Ma, Sean and Huang, Zhiheng and Karpathy, Andrej and Khosla, Aditya and Bernstein, Michael and others},
  journal={International journal of computer vision},
  volume={115},
  pages={211--252},
  year={2015},
  publisher={Springer}
}

@inproceedings{collab,
  title={Deep graph kernels},
  author={Yanardag, Pinar and Vishwanathan, SVN},
  booktitle={Proceedings of the 21th ACM SIGKDD international conference on knowledge discovery and data mining},
  pages={1365--1374},
  year={2015}
}

@article{tudataset,
  title={Tudataset: A collection of benchmark datasets for learning with graphs},
  author={Morris, Christopher and Kriege, Nils M and Bause, Franka and Kersting, Kristian and Mutzel, Petra and Neumann, Marion},
  journal={arXiv preprint arXiv:2007.08663},
  year={2020}
}

@article{nciprotein,
  title={Protein function prediction via graph kernels},
  author={Borgwardt, Karsten M and Ong, Cheng Soon and Sch{\"o}nauer, Stefan and Vishwanathan, SVN and Smola, Alex J and Kriegel, Hans-Peter},
  journal={Bioinformatics},
  volume={21},
  number={suppl\_1},
  pages={i47--i56},
  year={2005},
  publisher={Oxford University Press}
}

@article{stsb,
  title={Semeval-2017 task 1: Semantic textual similarity-multilingual and cross-lingual focused evaluation},
  author={Cer, Daniel and Diab, Mona and Agirre, Eneko and Lopez-Gazpio, Inigo and Specia, Lucia},
  journal={arXiv preprint arXiv:1708.00055},
  year={2017}
}

@inproceedings{sickr,
  title={Semeval-2014 task 1: Evaluation of compositional distributional semantic models on full sentences through semantic relatedness and textual entailment},
  author={Marelli, Marco and Bentivogli, Luisa and Baroni, Marco and Bernardi, Raffaella and Menini, Stefano and Zamparelli, Roberto},
  booktitle={Proceedings of the 8th international workshop on semantic evaluation (SemEval 2014)},
  pages={1--8},
  year={2014}
}

@inproceedings{resnet,
  title={Deep residual learning for image recognition},
  author={He, Kaiming and Zhang, Xiangyu and Ren, Shaoqing and Sun, Jian},
  booktitle={Proceedings of the IEEE conference on computer vision and pattern recognition},
  pages={770--778},
  year={2016}
}

@inproceedings{bert,
  title={Bert: Pre-training of deep bidirectional transformers for language understanding},
  author={Devlin, Jacob and Chang, Ming-Wei and Lee, Kenton and Toutanova, Kristina},
  booktitle={Proceedings of the 2019 conference of the North American chapter of the association for computational linguistics: human language technologies, volume 1 (long and short papers)},
  pages={4171--4186},
  year={2019}
}

@article{scl,
  title={Supervised contrastive learning},
  author={Khosla, Prannay and Teterwak, Piotr and Wang, Chen and Sarna, Aaron and Tian, Yonglong and Isola, Phillip and Maschinot, Aaron and Liu, Ce and Krishnan, Dilip},
  journal={Advances in neural information processing systems},
  volume={33},
  pages={18661--18673},
  year={2020}
}

@article{simcse++,
  title={SimCSE++: Improving contrastive learning for sentence embeddings from two perspectives},
  author={Xu, Jiahao and Shao, Wei and Chen, Lihui and Liu, Lemao},
  journal={arXiv preprint arXiv:2305.13192},
  year={2023}
}

@article{cert,
  title={Cert: Contrastive self-supervised learning for language understanding},
  author={Fang, Hongchao and Wang, Sicheng and Zhou, Meng and Ding, Jiayuan and Xie, Pengtao},
  journal={arXiv preprint arXiv:2005.12766},
  year={2020}
}

@article{consert,
  title={Consert: A contrastive framework for self-supervised sentence representation transfer},
  author={Yan, Yuanmeng and Li, Rumei and Wang, Sirui and Zhang, Fuzheng and Wu, Wei and Xu, Weiran},
  journal={arXiv preprint arXiv:2105.11741},
  year={2021}
}

@misc{signclip,
      title={SignCLIP: Connecting Text and Sign Language by Contrastive Learning}, 
      author={Zifan Jiang and Gerard Sant and Amit Moryossef and Mathias Müller and Rico Sennrich and Sarah Ebling},
      year={2024},
      eprint={2407.01264},
      archivePrefix={arXiv},
      primaryClass={cs.CL},
      url={https://arxiv.org/abs/2407.01264}, 
}

@inproceedings{joao,
  title={Graph contrastive learning automated},
  author={You, Yuning and Chen, Tianlong and Shen, Yang and Wang, Zhangyang},
  booktitle={International conference on machine learning},
  pages={12121--12132},
  year={2021},
  organization={PMLR}
}

@inproceedings{ref1,
  title={Contrastive multiview coding},
  author={Tian, Yonglong and Krishnan, Dilip and Isola, Phillip},
  booktitle={Computer Vision--ECCV 2020: 16th European Conference, Glasgow, UK, August 23--28, 2020, Proceedings, Part XI 16},
  pages={776--794},
  year={2020},
  organization={Springer}
}

@inproceedings{ref2,
  title={Contrastive learning with synthetic positives},
  author={Zeng, Dewen and Wu, Yawen and Hu, Xinrong and Xu, Xiaowei and Shi, Yiyu},
  booktitle={European Conference on Computer Vision},
  pages={430--447},
  year={2024},
  organization={Springer}
}

@inproceedings{ref3,
  title={How Does SimSiam Avoid Collapse Without Negative Samples? A Unified Understanding with Self-supervised Contrastive Learning},
  author={Zhang, Chaoning and Zhang, Kang and Zhang, Chenshuang and Pham, Trung X and Yoo, Chang D and Kweon, In So},
  booktitle={International Conference on Learning Representations},
  year={2022}
}

@article{ref4,
  title={A Survey on Self-supervised Contrastive Learning for Multimodal Text-Image Analysis},
  author={Khan, Asifullah and Asmatullah, Laiba and Malik, Anza and Khan, Shahzaib and Asif, Hamna},
  journal={arXiv preprint arXiv:2503.11101},
  year={2025}
}

@article{ref5,
  title={LOHA: Direct Graph Spectral Contrastive Learning Between Low-pass and High-pass Views},
  author={Zou, Ziyun and Jiang, Yinghui and Shen, Lian and Liu, Juan and Liu, Xiangrong},
  journal={arXiv preprint arXiv:2501.02969},
  year={2025}
}

@article{ref6,
  title={A survey on image data augmentation for deep learning},
  author={Shorten, Connor and Khoshgoftaar, Taghi M},
  journal={Journal of big data},
  volume={6},
  number={1},
  pages={1--48},
  year={2019},
  publisher={Springer}
}

@article{ref7,
  title={A survey on data augmentation for text classification},
  author={Bayer, Markus and Kaufhold, Marc-Andr{\'e} and Reuter, Christian},
  journal={ACM Computing Surveys},
  volume={55},
  number={7},
  pages={1--39},
  year={2022},
  publisher={ACM New York, NY}
}

@article{ref8,
  title={Data augmentation on graphs: a technical survey},
  author={Zhou, Jiajun and Xie, Chenxuan and Gong, Shengbo and Wen, Zhenyu and Zhao, Xiangyu and Xuan, Qi and Yang, Xiaoniu},
  journal={ACM Computing Surveys},
  year={2023},
  publisher={ACM New York, NY}
}

@article{ref9,
  title={Simple and asymmetric graph contrastive learning without augmentations},
  author={Xiao, Teng and Zhu, Huaisheng and Chen, Zhengyu and Wang, Suhang},
  journal={Advances in neural information processing systems},
  volume={36},
  pages={16129--16152},
  year={2023}
}

@inproceedings{ref10,
  title={Simgrace: A simple framework for graph contrastive learning without data augmentation},
  author={Xia, Jun and Wu, Lirong and Chen, Jintao and Hu, Bozhen and Li, Stan Z},
  booktitle={Proceedings of the ACM web conference 2022},
  pages={1070--1079},
  year={2022}
}

@article{1992,
  title={Self-organizing neural network that discovers surfaces in random-dot stereograms},
  author={Becker, Suzanna and Hinton, Geoffrey E},
  journal={Nature},
  volume={355},
  number={6356},
  pages={161--163},
  year={1992},
  publisher={Nature Publishing Group UK London}
}

@inproceedings{2005,
  title={Learning a similarity metric discriminatively, with application to face verification},
  author={Chopra, Sumit and Hadsell, Raia and LeCun, Yann},
  booktitle={2005 IEEE computer society conference on computer vision and pattern recognition (CVPR'05)},
  volume={1},
  pages={539--546},
  year={2005},
  organization={IEEE}
}

@inproceedings{2006,
  title={Dimensionality reduction by learning an invariant mapping},
  author={Hadsell, Raia and Chopra, Sumit and LeCun, Yann},
  booktitle={2006 IEEE computer society conference on computer vision and pattern recognition (CVPR'06)},
  volume={2},
  pages={1735--1742},
  year={2006},
  organization={IEEE}
}

@article{negcl,
  title={Hard negative mixing for contrastive learning},
  author={Kalantidis, Yannis and Sariyildiz, Mert Bulent and Pion, Noe and Weinzaepfel, Philippe and Larlus, Diane},
  journal={Advances in neural information processing systems},
  volume={33},
  pages={21798--21809},
  year={2020}
}

@article{negcl2,
  title={Contrastive learning with hard negative samples},
  author={Robinson, Joshua and Chuang, Ching-Yao and Sra, Suvrit and Jegelka, Stefanie},
  journal={arXiv preprint arXiv:2010.04592},
  year={2020}
}

@inproceedings{cmc,
  title={Contrastive multiview coding},
  author={Tian, Yonglong and Krishnan, Dilip and Isola, Phillip},
  booktitle={Computer Vision--ECCV 2020: 16th European Conference, Glasgow, UK, August 23--28, 2020, Proceedings, Part XI 16},
  pages={776--794},
  year={2020},
  organization={Springer}
}

@inproceedings{ref11,
  title={With a little help from my friends: Nearest-neighbor contrastive learning of visual representations},
  author={Dwibedi, Debidatta and Aytar, Yusuf and Tompson, Jonathan and Sermanet, Pierre and Zisserman, Andrew},
  booktitle={Proceedings of the IEEE/CVF international conference on computer vision},
  pages={9588--9597},
  year={2021}
}

@inproceedings{ref12,
  title={Timesurl: Self-supervised contrastive learning for universal time series representation learning},
  author={Liu, Jiexi and Chen, Songcan},
  booktitle={Proceedings of the AAAI conference on artificial intelligence},
  volume={38},
  number={12},
  pages={13918--13926},
  year={2024}
}

@article{2016,
  title={Improved deep metric learning with multi-class n-pair loss objective},
  author={Sohn, Kihyuk},
  journal={Advances in neural information processing systems},
  volume={29},
  year={2016}
}

@inproceedings{ref13,
  title={Contrastive multi-view representation learning on graphs},
  author={Hassani, Kaveh and Khasahmadi, Amir Hosein},
  booktitle={International conference on machine learning},
  pages={4116--4126},
  year={2020},
  organization={PMLR}
}

@article{ref14,
  title={Neural eigenfunctions are structured representation learners},
  author={Deng, Zhijie and Shi, Jiaxin and Zhang, Hao and Cui, Peng and Lu, Cewu and Zhu, Jun},
  journal={arXiv preprint arXiv:2210.12637},
  year={2022}
}

@inproceedings{ref15,
  title={Learning transferable visual models from natural language supervision},
  author={Radford, Alec and Kim, Jong Wook and Hallacy, Chris and Ramesh, Aditya and Goh, Gabriel and Agarwal, Sandhini and Sastry, Girish and Askell, Amanda and Mishkin, Pamela and Clark, Jack and others},
  booktitle={International conference on machine learning},
  pages={8748--8763},
  year={2021},
  organization={PmLR}
}

@article{collapse,
  title={Understanding dimensional collapse in contrastive self-supervised learning},
  author={Jing, Li and Vincent, Pascal and LeCun, Yann and Tian, Yuandong},
  journal={arXiv preprint arXiv:2110.09348},
  year={2021}
}

@article{ref16,
  title={Wasserstein dependency measure for representation learning},
  author={Ozair, Sherjil and Lynch, Corey and Bengio, Yoshua and Van den Oord, Aaron and Levine, Sergey and Sermanet, Pierre},
  journal={Advances in Neural Information Processing Systems},
  volume={32},
  year={2019}
}

@inproceedings{ref17,
  title={SELF-SUPERVISED REPRESENTATION LEARNING WITH RELATIVE PREDICTIVE CODING},
  author={Tsai, Yao Hung Hubert and Ma, Martin Q and Yang, Muqiao and Zhao, Han and Morency, Louis Philippe and Salakhutdinov, Ruslan},
  booktitle={9th International Conference on Learning Representations, ICLR 2021},
  year={2021}
}

@inproceedings{ref18,
  title={Learning deep representations by mutual information estimation and maximization},
  author={Hjelm, R Devon and Fedorov, Alex and Lavoie-Marchildon, Samuel and Grewal, Karan and Bachman, Phil and Trischler, Adam and Bengio, Yoshua},
  booktitle={International Conference on Learning Representations},
    year={2018}
}

@article{ref19,
  title={Improving fine-tuning of self-supervised models with contrastive initialization},
  author={Pan, Haolin and Guo, Yong and Deng, Qinyi and Yang, Haomin and Chen, Jian and Chen, Yiqun},
  journal={Neural Networks},
  volume={159},
  pages={198--207},
  year={2023},
  publisher={Elsevier}
}

% %% else use the following coding to input the bibitems directly in the
% %% TeX file.

% %% Refer following link for more details about bibliography and citations.
% %% https://en.wikibooks.org/wiki/LaTeX/Bibliography_Management


% %% For numbered reference style

%   \textit{\LaTeX: a document preparation system},
%   Addison Wesley, Massachusetts,

% \end{thebibliography}
\end{document}